\documentclass[reqno]{amsart}
\usepackage{kotex}

\usepackage{amsmath,amsthm,amssymb,amsfonts}
\usepackage{mathrsfs}
\usepackage[a4paper, margin=2.5cm]{geometry}

\usepackage{hyperref}

\usepackage{graphicx,xcolor}

\usepackage{tikz,ytableau}

\usepackage{algorithm}
\usepackage{algpseudocode}
\usepackage{booktabs}
\usepackage{mathtools}

\usepackage{tabularx,booktabs}
\usepackage{adjustbox} % preamble
\usepackage{float}

\usepackage{todonotes}

\newcolumntype{Y}{>{\raggedright\arraybackslash}X}
\usetikzlibrary{decorations.pathreplacing}
\ytableausetup{aligntableaux=bottom}
\numberwithin{equation}{section}

\newtheorem{thm}{Theorem}[section]
\newtheorem{lem}[thm]{Lemma}

\newtheorem{rem}[thm]{Remark}
\theoremstyle{definition}
\newtheorem{exam}[thm]{Example}
\newtheorem{defn}[thm]{Definition}

\title{Trace Regularity PINNs: Enforcing $\mathrm{H}^{\frac{1}{2}}(\partial \Omega)$  for Boundary Data}
\begin{document}

\author[D. Kim]{Doyoon Kim}
\address[D. Kim]{Department of Mathematics, Korea University, 145 Anam-ro, Seongbuk-gu, Seoul, 02841, Republic of Korea}
\email{\href{mailto:doyoon_kim@korea.ac.kr}{\nolinkurl{doyoon_kim@korea.ac.kr}}}
\thanks{This work was supported by the National Research Foundation of Korea (NRF) grant funded by the Korea government (MSIT) (RS-2025-16065192) and by a Korea University Grant.}
\thanks{The source code for the numerical experiments in this paper is available at https://github.com/JunbinSong/TRPINN}

\author[J. Song]{Junbin Song}
\address[J. Song]{Department of Mathematics, Korea University, 145 Anam-ro, Seongbuk-gu, Seoul, 02841, Republic of Korea}
\email{najbya@korea.ac.kr}

\maketitle

\begin{abstract}
We propose an enhanced physics-informed neural network (PINN), the Trace Regularity Physics-Informed Neural Network (TRPINN), which enforces the boundary loss in the Sobolev–Slobodeckij norm $\mathrm{H}^{\frac{1}{2}}(\partial\Omega)$, the correct trace space associated with $\mathrm{H}^1(\Omega)$.
We reduce computational cost by computing only the theoretically essential portion of the semi-norm and enhance convergence stability by avoiding denominator evaluations in the discretization.
By incorporating the exact $\mathrm{H}^{\frac{1}{2}}(\partial\Omega)$ norm, we show that the approximation converges to the true solution in the $\mathrm{H}^{1}(\Omega)$ sense, and, through Neural Tangent Kernel (NTK) analysis, we demonstrate that TRPINN can converge faster than standard PINNs.
Numerical experiments on the Laplace equation with highly oscillatory Dirichlet boundary conditions exhibit cases where TRPINN succeeds even when standard PINNs fail, and show performance improvements of one to three decimal digits.
\end{abstract}

\section{Introduction}

In complex physical and engineering problems where partial differential equations (PDEs) are difficult to solve analytically, \emph{physics-informed neural networks} (PINNs) have emerged as a flexible and simple framework that approximates solutions by embedding the differential operator residuals and boundary/initial conditions directly into the loss function.
Compared to mesh-based numerical solvers such as the finite element or finite difference methods, PINNs are mesh-independent, provide a unified objective for forward and inverse problems even with incomplete observations, and leverage automatic differentiation to compute higher-order derivatives with ease.
By jointly minimizing the \emph{interior residuals} and the \emph{boundary mismatches}, PINNs pursue a globally consistent solution, which makes them attractive for complicated geometries, noisy data, unknown coefficients, and multi-physics couplings.
The learned neural surrogate is a continuous object, which further facilitates post-processing, sensitivity analysis, and uncertainty quantification.

We consider the following elliptic partial differential equation in divergence form with Dirichlet boundary conditions
\begin{equation}
  \label{eq02}
  \left\{
    \begin{aligned}
    &\sum_{i,j=1}^dD_i(a^{ij}D_ju) = f \quad \text{in}\; \Omega,\\
    &u|_{\partial \Omega} = g \quad \text{on}\; \partial \Omega.
    \end{aligned}
  \right.
\end{equation}
In what follows, the solution space is taken to be $\mathrm{H}^{1}(\Omega)$, $f \in \mathrm{L}^2(\Omega)$, and $\partial\Omega$ is sufficiently smooth (Lipschitz is enough).
By the trace theorem (see, for instance, \cite{MR503903,MR2352844}), the boundary data \(g\) must belong to \(\mathrm{H}^{\frac{1}{2}}(\partial \Omega)\).
Note that when the dimension $d=2$, the semi-norm part of the norm of $\mathrm{H}^{\frac{1}{2}}(\partial \Omega)$ is defined by
\begin{equation}
\label{eq03}
[g]_{\mathrm{H}^{\frac{1}{2}}(\partial\Omega)}^2
= \int_{\partial\Omega}\int_{\partial\Omega}
\frac{\bigl|g(x)-g(y)\bigr|^2}{\lvert x-y\rvert^{2}}\,
\mathrm{d}S(x)\,\mathrm{d}S(y),
\end{equation}
where $\mathrm{d}S(\cdot)$ denotes the arc-length measure on the boundary curve $\partial\Omega$.

To keep computations simple, conventional PINN formulations typically adopt an $\mathrm{L}^{2}(\partial\Omega)$ boundary loss and define the total loss accordingly:
\[
\ell_{\mathrm{PINN}}
\,=\, N_1\| f_{NN} - f\|_{\mathrm{L}^2(\Omega)}^2 + N_2\|g_{NN}-g\|_{\mathrm{L}^2(\partial\Omega)}^2,
\]
where $f_{NN} := \sum_{i,j=1}^dD_i(a^{ij}D_ju_{NN})$, $g_{NN}:= u_{NN}|_{\partial \Omega}$, and $u_{NN}$ is a neural network.
From a theoretical standpoint, the boundary loss needs to control the $\mathrm{H}^{\frac{1}{2}}(\partial \Omega)$ norm; imposing only $\mathrm{L}^2(\partial\Omega)$ agreement—as in standard PINNs—may be insufficient to control the energy-space regularity of $u_{NN}$ as an approximation of the true solution.
Indeed, a conventional PINN may produce an approximate solution that satisfies a discrepant boundary condition (see \tablename~\ref{7VA}) when the boundary data $g$ are not enforced in $\mathrm{H}^{\frac{1}{2}}(\partial \Omega)$.
This, in turn, can lead the PINN to misinterpret the governing PDE and yield an approximation corresponding to a problem with an altered boundary condition.
Such boundary errors may propagate into the interior, causing the method to fail to converge to the true solution.

Although numerous deep-learning-based PDE solvers--such as variational PINN (vPINN), hard-constraint PINN, and weak adversarial network (WAN), etc.
\cite{kharazmi2019variational,toscano2025pinns,lu2021physics,ZANG2020109409}--have been proposed, most rely on an $\mathrm{L}^2$-based boundary loss.
Recently, \cite{zhou2025ssbepinnsobolevboundaryscheme} proposed a PINN that enforces the boundary regularity $\mathrm{H}^1(\partial\Omega)$.
For a solution $u \in \mathrm{H}^1(\Omega)$ to \eqref{eq02}, the boundary value $g$ of $u$ is guaranteed to belong only to $\mathrm{H}^{\frac{1}{2}}(\partial \Omega)$.
Thus, imposing a higher requirement such as $\mathrm{H}^1(\partial\Omega)$ may be appropriate for problems admitting smoother solutions--for instance, equations in non-divergence form--but may not be necessary in the general divergence-form setting.
In particular, when $g \in \mathrm{H}^{\frac{1}{2}}(\partial\Omega)$, the method in \cite{zhou2025ssbepinnsobolevboundaryscheme} requires derivative information $Dg$ on the boundary, which is not always well defined (see Example \ref{example_g}).
In \cite{liu2023deep}, the authors proposed a deep Ritz method with an adaptive quadrature rule (DRM-AQR) employing $\mathrm{H}^{\frac{1}{2}}(\partial \Omega)$ norm for the boundary data.
To realize the semi-norm part of $\mathrm{H}^{\frac{1}{2}}(\partial \Omega)$ norm when the dimension $d=2$ (see \eqref{eq03}), the authors of \cite{liu2023deep} discretized the integrand in \eqref{eq03}.
It is worth noting that this integrand contains a denominator, which can lead to numerical instability during computation if the denominator becomes very small.
In their implementation, when the denominator becomes zero, the authors replaced the original integrand with the derivative $Dg$, which may be unavailable if $g$ belongs only to $\mathrm{H}^{\frac{1}{2}}(\partial\Omega)$.
Concerning the integration region considered in \cite{liu2023deep} for the double integral in \eqref{eq03}, see Section~\ref{Discretization of Semi-Norm} and \figurename~\ref{int_bddomain}(b).

Since the semi-norm \eqref{eq03} involves a double integral and a denominator, its numerical computation can be nontrivial.
To overcome these numerical challenges and to incorporate the $\mathrm{H}^{\frac{1}{2}}(\partial \Omega)$ norm within a PINN framework, we propose a numerical scheme for computing the semi-norm \eqref{eq03} and introduce the Trace Regularity 
Physics-Informed Neural Network (TRPINN).
Note that throughout the paper, we consider $d=2$, so that $\Omega \subset \mathbb{R}^2$.
In our PINN framework, the loss function is based on
\[
\ell_{\mathrm{TRPINN}}
\,=\, N_1\| f_{NN} - f\|_{\mathrm{L}^2(\Omega)}^2 + N_2\|g_{NN}-g\|_{\mathrm{H}^{\frac{1}{2}}(\partial\Omega)}^2,
\]
where $\|\cdot\|_{\mathrm{H}^{\frac{1}{2}}(\partial\Omega)}$ is the sum of the $\mathrm{L}^2(\partial\Omega)$ norm and the semi-norm in \eqref{eq03}. 
A detailed theoretical formulation of the semi-norm is provided in Section~\ref{Theoretical Framework}.

To enforce the $\mathrm{H}^{\frac{1}{2}}(\partial \Omega)$ norm in the PINN framework--that is, in the proposed TRPINN--we realize the semi-norm \eqref{eq03} by computing its double integral only over the essential regions; see \eqref{eq01} together with \figurename~\ref{int_bddomain}.
We evaluate \eqref{eq03} (in fact, \eqref{eq01}) in the numerical integration process without any consideration of denominator terms.
This is made possible by applying the trapezoidal rule, appropriately adapted to \eqref{eq01} for the computation of the semi-norm, which allows the integrand that originally contains a denominator to naturally transform into a denominator-free form; the corresponding discretization scheme is described in Section~\ref{Discretization of Semi-Norm}.
Moreover, the computation is performed without requiring any information about $Dg$.
%Intuitively, whereas $\mathrm{L}^2(\partial\Omega)$ only enforces pointwise value agreement, $\mathrm{H}^{\frac{1}{2}}(\partial\Omega)$ simultaneously controls relative differences of the boundary values, thereby aligning the boundary regularity with the interior energy norm and improving consistency.
TRPINN can be implemented under the same conditions as the vanilla PINN with the $\mathrm{L}^2(\partial \Omega)$ norm, without a significant increase in computational cost.
Consequently, the overall training time remains comparable for the same number of iterations, and in some cases, training even converges faster due to early stopping in the L-BFGS optimizer (see \tablename~\ref{V=20_ms}).

Since the boundary loss of TRPINN is based on the $\mathrm{H}^{\frac{1}{2}}(\partial\Omega)$ norm, the convergence of the loss function to zero implies that the obtained approximation converges to the exact solution in the $\mathrm{H}^1(\Omega)$ norm.
We also describe the \emph{neural tangent kernel} (NTK) for analyzing neural networks \cite{jacot2018neural}.
The NTK characterizes the training dynamics, and its eigenvalues allow one to assess convergence rates at individual data points.
From the NTK perspective, \cite{wang2022and,rahaman2019spectral,cao2019towards,tancik2020fourier,basri2020frequency} analyzed PINNs and noted that, in regions of the domain where the target solution exhibits high-frequency content, learning proceeds more slowly.
From this viewpoint, we show that our proposed TRPINN yields larger eigenvalues, thereby demonstrating improved performance.

%We add a new term to the boundary loss to enforce the following \(\mathrm{H}^{\frac{1}{2}}\) semi-norm for $g$ defined on $\partial\Omega$, where $\Omega\subset \mathbb{R}^2$.
%We assume that the domain has a sufficiently smooth boundary, for instance, a Lipschitz boundary.
%Rather than discretizing the semi-norm directly via a Monte Carlo double integration—which incurs \(O(N^{2})\) cost as well as an unavoidable denominator computation that may produce huge errors—we evaluate only the theoretically essential subset of interactions and adopt a trapezoidal–rule–based discretization.
%This reduces the cost to \(O(N)\) and introduces negligible overhead relative to the total loss, including the interior loss.
%Moreover, by avoiding denominator computations during discretization, the stability is substantially improved.
The remainder of the paper is organized as follows.
Section~\ref{Theoretical Framework} introduces the notation, function spaces, associated norms--with an emphasis on the sufficiency of the essential part of the semi-norm--and the assumptions on the PDE used in the theoretical analysis.
Section~\ref{Discretization of Semi-Norm} describes the sampling and integration methods for TRPINN.
Section~\ref{Trace Regularity Physics-informed neural networks} reviews the vanilla PINN and presents the TRPINN method, highlighting its loss function.
Section~\ref{sec_convergence} establishes a theoretical error estimates; subsequently, Section~\ref{Neural Tangent Kernel} compares the vanilla PINN and TRPINN from the NTK perspective.
In Section~\ref{Numerical Experiments}, comparative experiments between PINN and TRPINN are presented.
Discussion and conclusion appear in Sections~\ref{discussion} and~\ref{conclusion}, respectively.

\section{Theoretical Framework}
\label{Theoretical Framework}

To analyze \textsc{TRPINN} rigorously, we first fix notation and recall the function spaces and norms used throughout.
In particular, we highlight boundary trace spaces, which underlie our loss design and the estimates employed in later sections.
This section was written with reference to \cite{krylov2024lectures,evans2022partial,MR2352844}.

\subsection{Function Spaces and Norms}

In addition to the usual $\mathrm{L}^p(\Omega)$, $1 \le p \le \infty$, we use the following:
\begin{itemize}
  %\item \(\mathrm{L}^2(\Omega)\): space of square‐integrable functions over \(\Omega\).
  %\item \(\mathrm{L}^{\infty}(\Omega)\) : space of measurable functions on \(\Omega\) that are essentially bounded.
  \item $\mathrm{H}^k(\Omega)$, $k=1,2$: Sobolev spaces of functions with weak derivatives up to $k$-th order in \(\mathrm{L}^2(\Omega)\).
  \item \(\mathrm{H}^{\frac{3}{2}}(\partial \Omega)\), \(\mathrm{H}^{\frac{1}{2}}(\partial \Omega)\): Sobolev-Slobodeckij spaces on the boundary, typically defined via interpolation theory or using local charts and partitions of unity.

\item $\mathrm{W}_\infty^1(\Omega)$ : Sobolev space of functions with weak derivatives up to first order in \(\mathrm{L}^{\infty}(\Omega)\).
\item $C^{0,1}$ boundary or Lipschitz boundary: Near every boundary point (after a suitable rotation/shift), the boundary is the graph of a Lipschitz function over a hyperplane.
\end{itemize}

The associated norms are given by
\begin{align*}
%&\|u\|_{\mathrm{L}^2(\Omega)}^2
%= \int_{\Omega} \lvert u(x)\rvert^2 \, dx,\\%[0.5em]
%&\|u\|_{\mathrm{L}^\infty(\Omega)}=\operatorname*{ess\,sup}_{x\in\Omega}|u(x)|,\\
&\|u\|_{\mathrm{H}^k(\Omega)}^2 = \sum_{|\alpha|\leq k}\|D^\alpha u\|_{\mathrm{L}^2(\Omega)}, \,\, k=1,2, 
\\%[0.5em]
&\|g\|_{\mathrm{H}^{\frac{1}{2}}(\partial \Omega)}^2
= \|g\|_{\mathrm{L}^2(\partial \Omega)}^2
+ [g]_{\mathrm{H}^{\frac{1}{2}}(\partial \Omega)}^2,
\\%[0.5em]
&\|g\|_{\mathrm{H}^{\frac{3}{2}}(\partial \Omega)}^2
= \|g\|_{\mathrm{L}^2(\partial \Omega)}^2 + \|Dg\|_{\mathrm{L}^2(\partial \Omega)}^2
+ [Dg]_{\mathrm{H}^{\frac{1}{2}}(\partial \Omega)}^2,\\
&\|u\|_{\mathrm{W}^1_{\infty}(\Omega)} = \sum_{|\alpha|\leq 1}\|D^\alpha u\|_{\mathrm{L}^\infty(\Omega)}.
\end{align*}

Here, for $\Omega \subset \mathbb{R}^d$, the semi-norm $[g]_{\mathrm{H}^{s}(\partial \Omega)}$ of $g$ defined on $\partial\Omega$ with $s \in (0,1)$ is defined by
\[ 
[g]_{\mathrm{H}^{s}(\partial \Omega)}^2 = \int_{\partial \Omega}\!\!\int_{\partial \Omega}
\frac{\lvert g(x) - g(y)\rvert^2}{\lvert x - y\rvert^{(d-1) + 2s}}
\,dS(x)\,dS(y).
 \]
When the dimension of the domain is $d = 2$ and $s = \frac{1}{2}$, we have \eqref{eq03}.

\begin{lem}
  \label{delta-semi-lem} 
  Let $\Omega \subset \mathbb{R}^2$.
  For any fixed $\delta >0 $, 
  \begin{equation}
    \label{delta-semi} 
    [g]_{\mathrm{H}^{\frac{1}{2}}(\partial \Omega)}^2 \approx \|g\|_{\mathrm{L}^2(\partial \Omega)}^2 + \int_{\partial \Omega}\!\!\int_{\partial \Omega} \frac{\lvert g(x) - g(y)\rvert^2}{\lvert x - y\rvert^2} 1_{|x-y|<\delta} \,dS(x)\,dS(y)
  \end{equation}
  for all $g \in \mathrm{H}^{\frac{1}{2}}(\partial \Omega)$, where the semi-norm equivalence depends on $\delta$.
\end{lem}
Note that the fact that the indicator function $1_{\lvert x - y \rvert < \delta}$ is multiplied into the integrand means that we do not have to evaluate the whole of \eqref{eq03}.
Therefore, in \eqref{eq01}, the \(\mathrm{L}^2\) norm is absorbed into the existing \(\mathrm{L}^2\) term of the loss function, and the second term \([u]_{\frac{1}{2},\delta}^2\) on the right-hand side of \eqref{delta-semi} is incorporated as an additional loss component during model training.
Here
\begin{equation}
  \label{eq01}
  [g]_{\frac{1}{2},\delta}^2 = \int_{\partial \Omega}\!\!\int_{\partial \Omega} \frac{\lvert g(x) - g(y)\rvert^2}{\lvert x - y\rvert^2} 1_{|x-y|<\delta} \,dS(x)\,dS(y).
\end{equation}

\subsection{Model Problem and Assumptions}

Consider the following elliptic partial differential equation in divergence form:
\begin{equation}
\label{divergence-form}
\left\{
\begin{aligned}
&\mathcal{L} u = f + \sum_{i=1}^{d} D_if_i \quad \text{in}\; \Omega, \\
&u = g \quad \text{on} \; \partial \Omega,
\end{aligned}
\right.
\end{equation}
where \(\Omega \subset \mathbb{R}^d\) is an open, bounded domain with a \(C^{0,1}\) (Lipschitz) boundary.
The forcing term is given by \(f, f_i \in \mathrm{L}^2(\Omega)\), and the boundary data \(g \in \mathrm{H}^{\frac{1}{2}}(\partial \Omega)\).
$\mathcal{L}$ is defined by
\begin{equation}
  \label{L-oper}
  \mathcal{L}u = \sum_{i,j=1}^dD_j(a^{ij}D_i u) + \sum_i b^iD_iu + cu,
\end{equation}
assume that $a^{ij}$ satisfies the ellipticity condition, i.e., for some $\theta > 0$,
\[
\sum_{i,j=1}^d a^{ij}(x)\xi_i\xi_j \ge \theta|\xi|^2 \quad \text{for a.e. $x \in \Omega$ and all $\xi \in \mathbb{R}^d$,}
\] 
and that $a^{ij}$ all belong to $\mathrm{W}^{1}_{\infty}(\Omega)$.
Although the coefficients $b_i$ and $c$ need not be zero for applicability, we assume $b_i=c=0$ for clarity of exposition.
In general, the coefficients $a^{ij}$ are only required to be bounded and to satisfy the ellipticity condition, without any regularity assumptions.
In the Appendix, however, we show that assuming $a^{ij} \in \mathrm{W}_\infty^1(\Omega)$ is sufficient when seeking approximate solutions using PINNs.
Also note that, in this paper, we consider the case $f_i = 0$;
the case of nonzero $f_i$ will be addressed in a subsequent paper, in which we aim to discuss how to efficiently treat divergence type equations in the PINN framework.

As is well known, under the assumptions mentioned above, the equation \eqref{divergence-form} has a unique solution $u \in \mathrm{H}^1(\Omega)$.
For $u \in \mathrm{H}^1(\Omega)$ with a sufficiently regular $\partial\Omega$, it is also well known that the trace of $u$, i.e., $u|_{\partial\Omega}$ belongs to $\mathrm{H}^{\frac{1}{2}}(\partial\Omega)$.
Therefore, it is necessary (and sufficient) to work with the $\mathrm{H}^{\frac{1}{2}}(\partial\Omega)$ norm when dealing with equations of the form \eqref{divergence-form} with $g \in \mathrm{H}^{\frac{1}{2}}(\partial\Omega)$.

One may consider, instead of $\mathcal{L} u$, the non-divergence type operator
\[
Lu = a^{ij} D_{ij} u + b_i D_i u + c u,
\]
with boundary condition $g \in \mathrm{H}^{\frac{3}{2}}(\partial\Omega)$, where solutions are to be sought in $\mathrm{H}^2(\Omega)$.
The same formulation and numerical approach extend naturally to this setting without essential modification, but with the boundary data in $\mathrm{H}^{\frac{3}{2}}(\partial\Omega)$ rather than in $\mathrm{H}^{\frac{1}{2}}(\partial\Omega)$, so that the semi-norm discretization (see Section \ref{Discretization of Semi-Norm} below) needs to be applied to $Dg$ instead of $g$.

\section{Discretization of Semi-Norm}
\label{Discretization of Semi-Norm}

In the context of PINNs, the standard practice for enforcing boundary conditions is to minimize the residual in the \(\mathrm{L}^2(\partial \Omega)\) norm, namely
\[
\|g_{NN} - g\|_{\mathrm{L}^2(\partial \Omega)}.
\]
%revise
Since computing the \(\mathrm{H}^{\frac{1}{2}}(\partial \Omega)\) norm via a Monte-Carlo integration may incur high computational cost and error, standard PINNs perform calculations using only the \(\mathrm{L}^2\) norm.

To address this limitation, we propose a discretization scheme for the \(\mathrm{H}^{\frac{1}{2}}(\partial \Omega)\) norm that enables a more rigorous and physically consistent enforcement of boundary conditions in the PINN framework.
By incorporating the fractional norm \eqref{eq01} into the loss function, we aim to improve the accuracy and physical fidelity of the learned solution, especially for problems where boundary regularity plays a crucial role.

We apply the idea of the single trapezoidal rule to evaluate \eqref{eq01}.
The single trapezoidal rule approximates the integral between two very close points $a,b\in\mathbb{R}$ as follows:
\[
\int_a^b h(x)\,dx \approx \frac{b-a}{2}\bigl(h(a)+h(b)\bigr).
\]
Since the semi-norm involves a double integral, we extend the trapezoidal rule to the double integral form as follows:
\[
\int_a^b\int_a^b h(x,y)\,dx\,dy \approx \frac{(b-a)^2}{2^2}(h(a,b) + h(b,a)).
\]
By taking $h(x,y) = \frac{|g(x) - g(y)|^2}{|x-y|^2}$, we have
\begin{align}
  \int_a^b\int_a^b \frac{|g(x) - g(y)|^2}{|x-y|^2} \,dx\,dy &\approx \frac{(b-a)^2}{2}(\frac{|g(b) - g(a)|^2}{|b-a|^2})\\[0.3em]
  \label{tra}
  & = \frac{1}{2}|g(b) - g(a)|^2.
\end{align}
Based on this, we discretize \eqref{eq01} as follows:
\begin{align}
  \label{eq04}
  [g]_{\frac{1}{2},\delta}^2 & \le \sum_{i=1}^{N}\int_{x_i}^{x^{i+1}}\int_{x_i}^{x^{i+1}} \frac{|g(x) - g(y)|^2}{|x-y|^2} dx dy + \sum_{i=1}^{N}\int_{x_i}^{x^{i+2}}\int_{x_i}^{x^{i+2}} \frac{|g(x) - g(y)|^2}{|x-y|^2} dx dy\\
  \label{eq05}
  &\approx \sum_{i=1}^{N}|g(x_{i+1})-g(x_i)|^2 + \sum_{i=1}^{N}|g(x_{i+2})-g(x_i)|^2,
\end{align}
where $x_1,x_2,\dots,x_N$ are points sampled from the boundary $\partial\Omega$, and $x_{N+1} = x_1, x_{N+2} = x_2$.
Here, we assume that their ordering is determined by a fixed rule (e.g., counterclockwise if the boundary is a circle).
The above integrals in \eqref{eq04} are to be intepreted as line integrals over \( \partial \Omega \).
The number $\delta >0$ in \eqref{eq01} can be chosen sufficiently small with respect to the given sampling points.
The second terms in \eqref{eq04} and \eqref{eq05} are added to more effectively cover the support of the indicator function $1_{\lvert x-y\rvert<\delta}$.
The constant $\frac{1}{2}$ in \eqref{tra} has been omitted.
Accordingly, we define the discretized semi-norm as follows:
\begin{equation}
  \label{eq06}
  [g]_{\frac{1}{2}}^2 = \sum_{i=1}^{N}|g(x_{i+1})-g(x_i)|^2 + \sum_{i=1}^{N}|g(x_{i+2})-g(x_i)|^2.
\end{equation}

Lemma~\ref{delta-semi-lem} implies that, instead of evaluating the semi-norm over the entire region, we may restrict the integration region to the support of the indicator function \(1_{|x-y|<\delta}\), which substantially reduces the computational cost. The complexity of our method scales linearly with the number \(N\) of sampled boundary points, i.e., \(O(N)\) (see~\eqref{eq06}). By contrast, evaluating the double integral in the semi-norm \eqref{eq03} over the full region via Monte Carlo integration would incur a cost of \(O(N^2)\). In both vanilla PINN and TRPINN, most of the computational burden arises from the interior residuals; therefore, the modest increase at the boundary has negligible impact on the overall training time. Empirically, Table~\ref{V=20_ms} shows no material difference. Moreover, in \eqref{tra} the denominator term cancels out, so TRPINN evaluates the semi-norm without computing the denominator, which improves numerical stability.

\begin{figure}[H]
    \centering
    \includegraphics[scale=0.3]{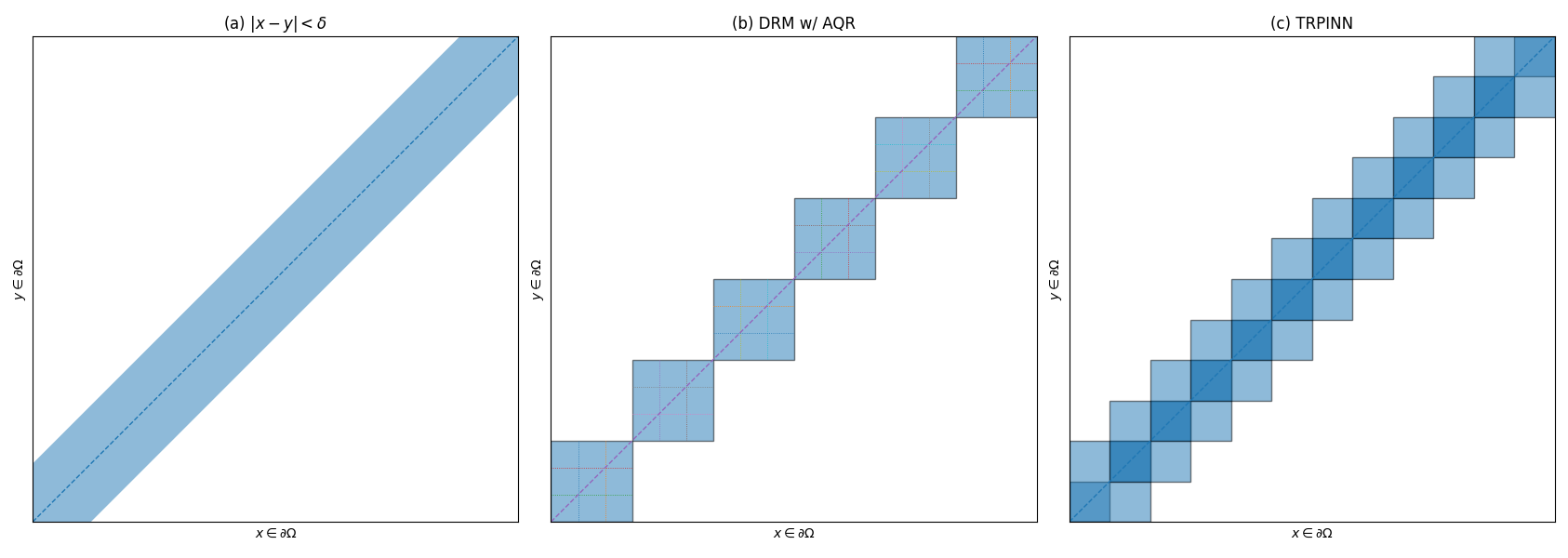}
    \caption{The integration regions for (a)–(c) correspond to the two-dimensional case ($d = 2$).
    (a) The integration region corresponding to Eq.~\eqref{eq01}.
    (b) The integration region used for the numerical integration of the boundary loss in the DRM with AQR method proposed in \cite{liu2023deep}, where the squares composed of 9 small cells correspond to $\mathcal{E}_D$, and the small cells correspond to $\mathcal{E}_D^k$ (see \cite{liu2023deep}).
    (c) The integration region for the numerical integration of the boundary loss in the proposed TRPINN method.}
    \label{int_bddomain}
\end{figure}

\figurename~\ref{int_bddomain} visually demonstrates the numerical integration regions for DRM-AQR and TRPINN (see panels~(b) and~(c), respectively).
Both methods avoid performing computations over the entire region $\partial \Omega \times \partial \Omega$ to reduce the computational cost.
Panel~(a) illustrates the integration region of \eqref{eq01}, and panel~(c) demonstrates that TRPINN can fully cover this region, whereas in panel~(b), part of the region $|x-y|<\delta$ remains uncovered

%gaps arise between the squares so that some regions remain uncovered.

\section{Trace Regularity Physics-informed neural networks}
\label{Trace Regularity Physics-informed neural networks}

\subsection{Conventional PINN}

One can find the general framework and terminology of physics-informed neural networks (PINNs) in \cite{raissi2017physics,raissi2019physics,toscano2025pinns}.
PINN is a neural network trained using the following loss:
\begin{align}
\label{PINN_loss}
\ell_{PINN} = \alpha \,\ell_{\mathrm{inside}} + \beta \,\ell_{\mathrm{boundary}},
\end{align}
where 
\begin{align*}
&\ell_{\mathrm{inside}}(\theta) = \lVert f_{NN}(\cdot, \theta) - f\rVert_{\ell^2(\Omega)}^2 \approx \frac{1}{N_f}\sum_{i=1}^{N_f} \rvert f_{NN}(x_i^{in}, \theta)-f(x_i^{in})\lvert^2, \\
&\ell_{\mathrm{boundary}}(\theta)=\lVert g_{NN}(\cdot, \theta)\bigr. - g \rVert_{\ell^2(\partial \Omega)}^2 \approx \frac{1}{N_g} \sum_{i=1}^{N_g} \rvert g_{NN}(x_i^{bd}, \theta) - g(x_i^{bd}) \lvert^2.
\end{align*}
Here, $f_{NN} := \sum_{i,j=1}^{d}D_i(a^{ij}D_ju_{NN})$, $g_{NN}:= u_{NN}|_{\partial \Omega}$, and $u_{NN}$ denotes a deep neural network (multi‐layer perceptron) that approximates the solution $u$, and $\theta$ represents the trainable parameters of $u_{\mathrm{NN}}$.
The points $x_i^{\mathrm{in}}$ are collocation points in the interior domain $\Omega$, with $N_f$ denoting their total number, while $x_i^{\mathrm{bd}}$ are boundary points of $\Omega$, and $N_g$ denotes their total number.
$\alpha$ and $\beta$ represent weights for the inside and boundary loss, respectively, and can be adjusted according to the specifics of the problem.
Recent studies have also introduced adaptive weighting methods to dynamically determine these parameters \cite{wang2022and}.
The computation of \(f_{NN}\) was carried out using the automatic differentiation capabilities of various frameworks (e.g., PyTorch, TensorFlow).

%Conventional PINNs, as discussed above, may exhibit degraded convergence in the interior due to simplifications of the boundary loss and are particularly vulnerable when approximating solutions with high-frequency components \cite{tancik2020fourier}.
%To overcome this limitation, we propose TRPINN in the following section.

Conventional PINNs often suffer from sensitivity to the loss weights $\alpha$ and $\beta$, which can bias training toward boundary data, underfit the interior residual, and slow or destabilize convergence---particularly for solutions with pronounced high-frequency components.
Although adaptive weighting strategies have been explored to alleviate these issues \cite{mcclenny2023self, wang2024improved, wang2022and}, they still rely on the $\mathrm{L}^2$ boundary loss, which may not capture the exact regularity of the boundary data.
To address these limitations, we propose TRPINN, which is designed to enhance convergence and to be substantially more robust to the choice of loss weights.

\subsection{Trace Regularity PINN (TRPINN)}

Our proposed trace-regularity physics-informed neural network (TRPINN) is trained using the following loss function:
\begin{equation}
\label{TRPINN_loss}
\ell_{TRPINN} = \alpha \,\ell_{\mathrm{inside}} + \beta \,\ell_{\mathrm{boundary}} + \gamma \ell_{\mathrm{semi}},
\end{equation}
where
\[
\ell_{\mathrm{semi}}(\theta)
:= [g_{NN}(\cdot, \theta) - g]_{\frac{1}{2}}^2 .
\]
Here, $\ell_{\mathrm{inside}}$ , $\ell_{\mathrm{boundary}}$, $\alpha$, $\beta$, and $\gamma$ are as defined earlier and $[\cdot]_{\frac{1}{2}}$ is the discretized semi-norm in \eqref{eq06}.

We improve performance by augmenting the loss with a semi-norm, thereby enforcing the theoretically exact norm.
In the next two sections, drawing on PDE theory and an analysis of the neural tangent kernel (NTK), we show that strengthening the boundary loss promotes convergence to accurate solutions and mitigates vulnerability to high-frequency components.

In addition, we emphasize that, by restricting computations to only those theoretically necessary, the computational cost remains comparable to that of the standard PINN, and that avoiding denominator evaluations in the computation process improves the stability of the loss calculation.
Our numerical examples in Section~\ref{Numerical Experiments} show that, in cases where the boundary exhibits high-frequency components, the proposed approach achieves substantial improvements over the standard PINN.

\section{\(\mathrm{H}^1\)-convergence in the interior $\Omega$}
\label{sec_convergence}

This section presents theoretical analysis demonstrating that our proposed \textsc{TRPINN} guarantees convergence to a more accurate solution by enforcing the appropriate norm.

\begin{thm}
  Let $\Omega$ be a $C^{0,1}$ domain and $a^{ij} \in \mathrm{W}_\infty^1(\Omega)$
  Then for all $u \in \mathrm{H}^1(\Omega)$,
  \begin{equation}
    \label{estimate}
    \lVert u\rVert_{\mathrm{H}^1(\Omega)} \le C \lVert f \rVert_{\mathrm{L}^2(\Omega)} + C \lVert g \rVert_{\mathrm{H}^{\frac{1}{2}}(\partial \Omega)},
  \end{equation}
  where $f = \sum_{i,j=1}^{d}D_i(a^{ij}D_ju)$, $g = u|_{\partial\Omega}$ and $C>0$ is independent of $u$.
\end{thm}
This theorem is well known; see, for instance, \cite{evans2022partial} for a proof.
\begin{rem}
  Let \(u\) be the ground-truth solution to \eqref{divergence-form} and $u_{NN}$ be a neural network.
  Then $u - u_{NN}$ is a solution to the following equation:
  \begin{equation}
  \left\{
  \begin{aligned}
    &\mathcal{L}(u-u_{NN}) = f - f_{NN} \quad \text{in}\; \Omega, \\
    &(u - u_{NN})|_{\partial\Omega} = g - g_{NN} \quad \text{on}\; \partial\Omega,
  \end{aligned}
  \right.
  \end{equation}
where $f_{NN} = \sum_{i,j=1}^{d}D_i(a^{ij}D_ju_{NN})$, $g_{NN} = u_{NN}|_{\partial \Omega}$.
Then, by \eqref{estimate}, we have
  \begin{equation}
    \label{Hconv}
    \lVert u - u_{NN}\rVert_{\mathrm{H}^1(\Omega)} \le C \lVert f - f_{NN}\rVert_{\mathrm{L}^2(\Omega)} + C \lVert g - g_{NN} \rVert_{\mathrm{H}^{\frac{1}{2}}(\partial \Omega)}.
  \end{equation}
Therefore, since TRPINN enforces RHS of \eqref{Hconv}, if the loss converges to zero, the approximate solution \(u_{\mathrm{NN}}\) is guaranteed to converge to the ground-truth solution \(u\) in the \(\mathrm{H}^1(\Omega)\)-norm.
\end{rem}

Conventional PINNs enforces only the $\mathrm{L}^2(\partial \Omega)$ norm on the boundary; therefore, even if the loss converges to zero, \(\mathrm{H}^1(\Omega)\) convergence in the interior is not guaranteed.
\cite{zhou2025ssbepinnsobolevboundaryscheme} presents examples showing that, when only $\mathrm{L}^{2}(\partial\Omega)$ is enforced, conventional PINNs can fail to achieve $\mathrm{H}^{1}$ convergence in the interior of the domain $\Omega$.

On the other hand, for equations in divergence form, imposing an unnecessarily strong boundary condition in PINNs can cause issues.
Indeed, by the trace theorem, the boundary condition $g$ satisfies \(g \in \mathrm{H}^{\frac{1}{2}}(\partial \Omega)\).
Thus, if \(g \in \mathrm{H}^{\frac{1}{2}}(\partial \Omega)\) but \(g \notin \mathrm{H}^{1}(\partial \Omega)\) (see Example~\ref{example_g} below), a PINN that penalizes the $\mathrm{H}^1(\partial\Omega)$-boundary norm in its loss cannot properly handle such a boundary condition $g$.

\begin{exam}
  \label{example_g}
  Define 
  \[
    g_0(t) = |t|^{\frac{1}{2}}
  \]
  for $t \in [-\pi,\pi]$.
  Since $g_0(-\pi) = g_0(\pi)$, we may regard it as a function on $\partial B$.
  Then clearly $\rVert g_0 \lVert_{\mathrm{L}^2(\partial B)} < \infty$ and we have
  \[
    [ g_0 ]_{\mathrm{H}^{\frac{1}{2}}(\partial B)}^2 \approx \int_{-\pi}^{\pi}\int_{-\pi}^{\pi} \frac{|g_0(t_1) - g_0(t_2)|^2}{|t_1-t_2|^2} \; dt_1 dt_2 = 12\pi \log2 - 2\pi^2,
  \]
  so that $\rVert g_0 \lVert_{\mathrm{H}^{\frac{1}{2}}(\partial B)} < \infty$.
  However,
  \[
    \rVert Dg_0 \lVert_{\mathrm{L}^2(\partial B)}^2 \approx \int_{-\pi}^{\pi}\frac{1}{4|t|} dt = \infty,
  \]
  which shows that $\rVert g_0 \lVert_{\mathrm{H}^1(\partial B)} = \infty$.
  Define a periodic function $g_1$ by 
  \begin{align*}
    g_1(t) = \begin{cases}
      \frac{1}{2} - g_0(\frac{t-\pi/2}{2\pi}) \quad \quad &0 \le t < \pi, \\
      g_0(\frac{t-3\pi/2}{2\pi}) - \frac{1}{2} \quad \quad &\pi \le t \le 2\pi.
    \end{cases}
  \end{align*}
%We can control $[g_1]_{\mathrm{H}^{\frac{1}{2}}(\partial B)}$ by $[g_0]_{\mathrm{H}^{\frac{1}{2}}(\partial B)}$ and by the second-order Taylor expansion of $g_0\!\left(\frac{t-\pi/2}{2\pi}\right)$ at $0$, which yields finiteness, i.e.
One can check that
\[
\rVert g_1 \lVert_{\mathrm{H}^{\frac{1}{2}}(\partial B)}<\infty \quad \text{and} \quad \|Dg_1\|_{\mathrm{L}^2(\partial B)}=\infty,
\]
so that $\|g_1\|_{\mathrm{H}^{2}(\partial B)}=\infty.$
If the boundary condition $g$ is given by
\begin{equation}
  \label{VA}
  g(t)=\frac{1}{V}\,g_1(Vt),  
\end{equation}
where $V$ is a positive integer representing the oscillation frequency, then $\|g\|_{\mathrm{H}^{\frac{1}{2}}(\partial B)}$ is finite while $\|g\|_{\mathrm{H}^{1}(\partial B)}$ is infinite.
Therefore, the $\mathrm{H}^{1}(\partial B)$ regularity cannot be enforced; only the $\mathrm{H}^{\frac{1}{2}}(\partial B)$ regularity can be enforced.
In our experiments we observe that even with $V=7$ the vanilla PINN completely fails to converge.

\end{exam}

 %This shows that $||Du_{NN} - Dg_0||_{\mathrm{L}^2(\partial \Omega)}$ is not computable. Therefore, enforcing the $\mathrm{H}^1(\partial \Omega)$ norm on the boundary may be impractical.

The auxiliary data \(Dg\) may be unavailable.
Even when such information is accessible, incorporating it requires additional auto-differentiations of the neural network \(u_{\mathrm{NN}}\), thereby increasing the computational cost.
In contrast, the proposed TRPINN does not require information about \(Dg\).
As previously explained, by discretizing the \(\mathrm{H}^{\frac{1}{2}}(\partial\Omega)\)-norm via the semi-norm \([g_{NN}-g]_{\frac{1}{2},\delta}\) (see \eqref{eq06}), the overall computational cost scarcely increases and thus remains comparable to that of a standard PINN.

\section{Neural Tangent Kernel}
\label{Neural Tangent Kernel}

This section introduces the Neural Tangent Kernel (NTK) to provide theoretical support for the performance of TRPINN.
The NTK has proved useful for analyzing the dynamics of neural network training \cite{jacot2018neural}, and prior work, for instance, \cite{wang2022and} has defined and analyzed NTK for PINNs.
Building on this line of work, we formalize the gradient flow of TRPINN from the NTK perspective and analyze it theoretically.

\begin{defn}
  \begin{align*}
    K(t) &\coloneqq
    \begin{bmatrix}
      K_{bb}(t) & K_{br}(t) \\
      K_{rb}(t) & K_{rr}(t)
    \end{bmatrix}.
  \end{align*}
  
\noindent
We call $K(t)$ the \emph{neural tangent kernel} (NTK) of a neural network $u_{NN}$ with respect to the differential operator $\mathcal{L} = \Delta$, where 
\begin{align*}
\big(K_{bb}\big)_{ij}(t)
&=
\left\langle 
  \nabla_{\theta}u_{NN}(x^b_{i};\,\theta(t))\,,
  \nabla_{\theta}u_{NN}(x^b_{j};\,\theta(t))
\right\rangle,\\[4pt]
\big(K_{br}\big)_{ij}(t)
&=
\left\langle
  \nabla_{\theta}u_{NN}(x^b_{i};\,\theta(t))\,,
  \nabla_{\theta}f_{NN}(x^b_{j};\,\theta(t))
\right\rangle,\\[4pt]
\big(K_{rr}\big)_{ij}(t)
&=
\left\langle
  \nabla_{\theta}f_{NN}(x^b_{i};\,\theta(t))\,,
  \nabla_{\theta}f_{NN}(x^b_{j};\,\theta(t))
\right\rangle, 
\end{align*}
$\big(K_{rb}\big) = \big(K_{br}\big)^{T}$.
$K_{bb}(t)\in\mathbb{R}^{N_b\times N_b}$,
$K_{br}(t)\in\mathbb{R}^{N_b\times N_r}$, and
$K_{rr}(t)\in\mathbb{R}^{N_r\times N_r}$.
$\left\langle\cdot,\cdot\right\rangle$ is the inner product with respect to $\theta$.
\end{defn}

\begin{rem}
  It was proved in \cite{jacot2018neural} that the NTK converges to a
  deterministic kernel $K^\ast$ as the network width tends to infinity, and that,
  under suitable conditions, it remains nearly constant throughout training.
  Consequently, the initial kernel $K(0)$ suffices to characterize the network’s
  training dynamics.
  We have
  \[
    K(0) \approx K(t) \approx K^*, \quad \forall t > 0.
  \]
\end{rem}

\begin{rem}
  Let $J_b = (\partial_{\theta_j} u_{NN}(x^b_i; \theta))_{ij}$ and $J_r = (\partial_{\theta_j} f_{NN}(x^r_i; \theta))_{ij}$ be the Jacobian matrices of $u_{NN}$ and $f_{NN}$, respectively.
  We have
  \begin{align*}
    K(t) = \begin{bmatrix}
  J_b(t)\\
  J_r(t)
\end{bmatrix}
\begin{bmatrix}
  J_b^{\top}(t) & J_r^{\top}(t)
\end{bmatrix}.
  \end{align*}
Note that $K(t)$ is positive semi-definite.
We employ the hyperbolic tangent activation function \(\tanh\).
It was proved in~\cite{carvalho2025positivity} that the kernel \(K(t)\) is positive definite for non-polynomial activation functions, so that it have spectrum decomposition.
\end{rem}

\subsection{Comparative analysis of PINN and TRPINN from the NTK viewpoint}

This subsection develops an NTK-based analysis of the training dynamics of PINN and TRPINN.

\begin{lem}
  Let $\theta^{PINN}, \theta^{TRPINN}$ be trainable parameters of PINN and TRPINN, respectively.
  Then we have
  \begin{align}
    \label{dynamics_PINN}
    \begin{bmatrix}
    \frac{d\,u_{NN}(x^b;\theta^{PINN}(t))}{dt} \\
    \frac{d\,f_{NN}(x^r;\theta^{PINN}(t))}{dt}
    \end{bmatrix}
    =
    -K(t)
    \begin{bmatrix}
    \frac{2}{N_b} I_{N_b} & 0 \\
    0 & \frac{2}{N_r} I_{N_r}
    \end{bmatrix}
    \begin{bmatrix}
    u_{NN}(x^b;\theta^{PINN}(t)) - g(x^b) \\
    f_{NN}(x^r;\theta^{PINN}(t)) - f(x^r)
    \end{bmatrix}\,,\\
    \label{dynamics_TRPINN}
    \begin{bmatrix}
    \frac{d\,u_{NN}(x^b;\theta^{TRPINN}(t))}{dt} \\
    \frac{d\,f_{NN}(x^r;\theta^{TRPINN}(t))}{dt}
    \end{bmatrix}
    =
    -K(t)
    \begin{bmatrix}
    M & 0 \\
    0 & \frac{2}{N_r} I_{N_r}
    \end{bmatrix}
    \begin{bmatrix}
    u_{NN}(x^b;\theta^{TRPINN}(t)) - g(x^b) \\
    f_{NN}(x^r;\theta^{TRPINN}(t)) - f(x^r)
    \end{bmatrix},
  \end{align}
  where $M = (\frac{2}{N_b} + 4)I_{N_b} - 2(P+P^{-1})$ and $P$ is the permutation matrix: 
  \begin{align*}
    P =
    \begin{bmatrix}
    0&1&0&0&...&0 \\
    0&0&1&0&...&0 \\
    0&0&0&1&...&0 \\
    &&\vdots&&&   \\
    0&0&0&...&0&1\\
    1&0&0&...&0&0
    \end{bmatrix}_{N_b \times N_b}.
  \end{align*}
\end{lem}
Here, \eqref{dynamics_PINN} follows from \cite{wang2022and}, and \eqref{dynamics_TRPINN} can be obtained by a straightforward calculation analogous to \eqref{dynamics_PINN}

\begin{rem}
  \label{PINN_eigen}
  Let \(K = K(0)\).
  From \cite{wang2022and}, by \eqref{dynamics_PINN} and using the solution of the ODE system, we compute
  \begin{align*}
  \begin{bmatrix}
    u_{NN}(x^b;\theta^{PINN}(t)) - g(x^b) \\
    f_{NN}(x^r;\theta^{PINN}(t)) - f(x^r)
  \end{bmatrix}
  \approx - e^{-K_pt} 
  \begin{bmatrix}
    g(x^b) \\
    f(x^r)
  \end{bmatrix},
  \end{align*}
  where
  \begin{align*}
    K_p = K
    \begin{bmatrix}
      \frac{2}{N_b} I_{N_b} & 0 \\
      0 & \frac{2}{N_r} I_{N_r}
    \end{bmatrix}.
  \end{align*}
  Noting that $K_p$ has the spectral decomposition $K_p= Q^{\top}\Lambda_p Q$, we then obtain
  \begin{align}
    \label{velocity_pinn}
    \begin{bmatrix}
    u_{NN}(x^b;\theta^{PINN}(t)) - g(x^b) \\
    f_{NN}(x^r;\theta^{PINN}(t)) - f(x^r)
  \end{bmatrix}
  \approx - Q^{\top} e^{-\Lambda_p t} Q 
  \begin{bmatrix}
    g(x^b) \\
    f(x^r)
  \end{bmatrix}.
  \end{align}
\end{rem}
The passage of time \(t\) represents the progression of neural network training.
Thus, if \(t \to \infty\), the right-hand side of \eqref{velocity_pinn} converges to zero.
In this case, the left-hand side corresponds to
\begin{align*}
  \begin{bmatrix}
    u_{NN}(x^b;\theta^{PINN}(t))\\
    f_{NN}(x^r;\theta^{PINN}(t))
  \end{bmatrix}
  \approx
  \begin{bmatrix}
    g(x^b) \\
    f(x^r)
  \end{bmatrix}.
\end{align*}
As training progresses over time, the eigenvalues (i.e., the components) of \(\Lambda_p\) characterize how quickly the right-hand side converges to zero.
Therefore, the eigenvalues of \(\Lambda_p\) can serve as indicators of the convergence speed at each point \(x\).

\cite{wang2022and} reports that the matrix \(\Lambda_{p}\) possesses a large number of small eigenvalues and points out that training progresses slowly along directions associated with such small eigenvalues.
In addition, according to \cite{rahaman2019spectral}, PINNs struggle to learn solutions that exhibit strong oscillations.

We will show that our proposed TRPINN, which replaces the standard \(\mathrm{L}^{2}(\partial \Omega)\) discrepancy with the Sobolev–Slobodeckij norm \(\mathrm{H}^{\frac{1}{2}}(\partial \Omega)\), yields larger eigenvalues than a conventional PINN and achieves faster solution convergence, improving performance in oscillatory and ill-conditioned regimes.

\begin{rem}
 In \eqref{dynamics_TRPINN}, since \(M\) is a symmetric and positive definite matrix, 
\begin{align*}
  K_h = K
  \begin{bmatrix}
  M & 0 \\
  0 & \frac{2}{N_r} I_{N_r}
  \end{bmatrix}
\end{align*}
  also admits a spectral decomposition $K_h = Q'^{\top}\Lambda_h Q'$.
  As in the remark above, we can derive the following equation for TRPINN:
  \begin{align*}
  \begin{bmatrix}
    u_{NN}(x^b;\theta^{TRPINN}(t)) - g(x^b) \\
    f_{NN}(x^r;\theta^{TRPINN}(t)) - f(x^r)
  \end{bmatrix}
  \approx - Q'^{\top} e^{-\Lambda_h t} Q' 
  \begin{bmatrix}
    g(x^b) \\
    f(x^r)
  \end{bmatrix}.
  \end{align*}
\end{rem}
As $\Lambda_p$ in Remark \ref{PINN_eigen} represents the convergence rate of PINN, \(\Lambda_h\) represents that of TRPINN at each point \(x\).
The following lemma shows that TRPINN converges faster than the standard PINN.

\begin{lem}
  Let $\Lambda_p$ and $\Lambda_h$ be the matrices given above.
  Then
  \[
    (\Lambda_p)_i \leq (\Lambda_h)_i
  \]
  for each $i = 1, ... , N_b + N_r$.
\end{lem}
The proof of this lemma is elementary, so we omit it.

As discussed above, larger eigenvalues imply a higher convergence rate at each point.
Referring to \figurename~\ref{eigen}, we observe that the increase occurs primarily for small eigenvalues.
This shows that the speed disparity across points can be alleviated.

\subsection{Computed NTK Spectrum on boundary $\partial \Omega
$}
\begin{figure}[H]
    \centering
    \includegraphics[width=0.7\linewidth]{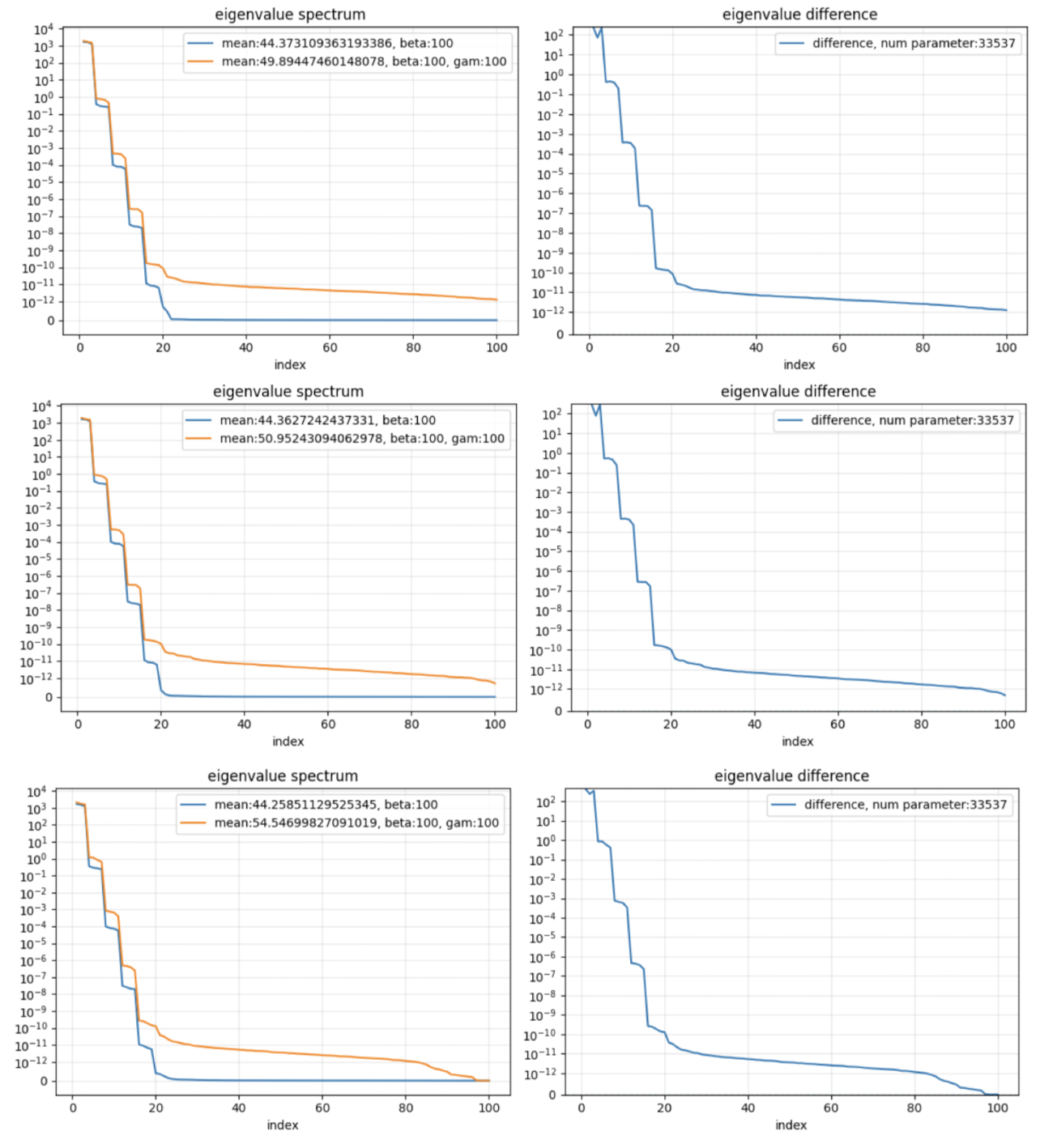} % 이미지 파일 이름
    \caption{The left column shows the magnitudes of the eigenvalues for PINN (blue line) and TRPINN (orange line), computed by extracting the largest one million.
    The right column shows the absolute value of their differences.
    From top to bottom, the boundary sampling methods tested are linspace, randomized, and uniform random.}
 % 캡션
    \label{eigen} % 참조용 라벨
\end{figure}

Computing $\Lambda_p$ and $\Lambda_h$ in their entirety is difficult.
To demonstrate this empirically, we computed and compared only the top-100 eigenvalues of $K_{bb}(1/N_b)I_{N_b}$ and $K_{bb}M$, i.e., the 100 largest values sorted in descending order.
We employed a neural network with a two-dimensional input and a scalar output.
The architecture comprised three fully connected hidden layers, each of width 128.
All loss functions were assigned identical scaling coefficients, \(\beta=\gamma=100\).
The rate \(\alpha\) was omitted because the objective is evaluated only on the boundary.
 \figurename~\ref{eigen} shows that the eigenvalues of $K_{bb}M$ are larger than those of $K_{bb}(1/N_b)I_{N_b}$.
 The left panel reports the top-100 eigenvalues computed separately, and the blue line corresponds to the eigenvalues of $K_{bb}(1/N_b)I_{N_b}$, whereas the orange line corresponds to those of $K_{bb}M$, and
 the right panel shows the difference obtained by subtracting the top-100 of $K_{bb}(1/N_b)I_{N_b}$ from the top-100 of $K_{bb}M$.
 From top to bottom, the rows correspond to the linspace sampling method, a randomized method, and a uniform random method, respectively.

Therefore, we have established—both theoretically and experimentally—that TRPINN exhibits larger eigenvalues.

\section{Numerical Experiments}
\label{Numerical Experiments}

In our numerical experiment, we consider the Laplace equation:
\begin{align}
  \label{exam-lap}
    \begin{cases}
        \Delta u = 0 \quad \text{in } \Omega,\\
        u|_{\partial\Omega} = g \quad \text{on }\partial \Omega.
    \end{cases}
\end{align}
The true solution is obtained via a harmonic extension computed using the Fast Fourier Transform (FFT).
The domain $\Omega$ is the unit ball (unit disk) in $\mathbb{R}^2$.
Unless otherwise specified, we used a neural network with three layers, each containing 128 units.
The rates of the loss functions, \(\alpha\), \(\beta\), and \(\gamma\), were set differently in each experiment.
For interior sampling, we used the uniform random method with 10,000 points, whereas for boundary sampling we used the randomized method \cite{wu2022randomized} with 201 points.
The activation function was \texttt{tanh}.
For optimization, we employed Adam $+$ L-BFGS, setting the maximum number of iterations to 50,000 for each.
We only consider cases in which an exact solution from harmonic extension is available in all experiments, so we can always compute the relative error during training.
Therefore, to compare learnability with the vanilla PINN, we report the lowest (best) relative \(\mathrm{H}^1(\Omega)\) error observed during training; in addition, we make use of the concurrently computed relative errors \(\mathrm{L}^2(\Omega)\), \(\mathrm{H}^{\frac{1}{2}}(\partial\Omega)\), and \(\mathrm{L}^2(\partial\Omega)\).
We compare the two models by varying the rates $\alpha$, $\beta$, and $\gamma$ in \eqref{PINN_loss} and \eqref{TRPINN_loss}.

\subsection{Boundary Condition with High Oscillations}

Our numerical experiment compares two models on learning true solutions that exhibit pronounced oscillations along the boundary, a regime in which standard PINNs are known to struggle.
Dirichlet boundary data given by
\begin{align*}
  g(x) = \sin(V \theta) \quad \text{on} \quad \partial\Omega,
\end{align*}
where $x=(x_1,x_2) = (r\cos\theta, r\sin\theta)$.

\begin{figure}[H]
    \centering
    \includegraphics[scale=0.25]{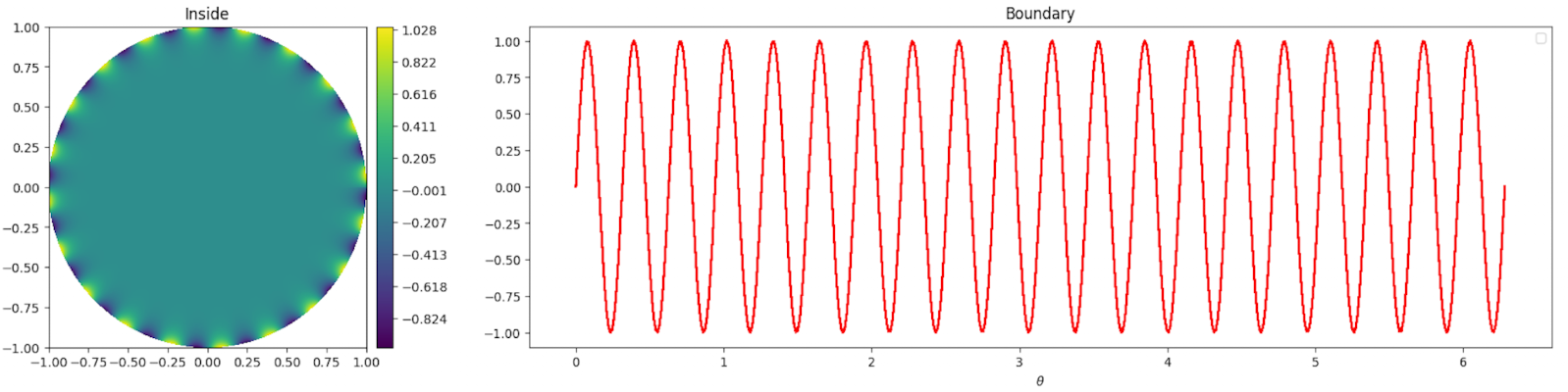} % 이미지 파일 이름
    \caption{Harmonic extension of $\sin(20\theta)$.
    This is used as the exact solution to compute the relative error.} % 캡션
    \label{20solution} % 참조용 라벨
\end{figure}

We consider the case \(V=20\).
In the conventional PINN, the loss comprises two terms, necessitating two weights, $\alpha_1$ and $\beta_1$.
In contrast, TRPINN involves three loss terms and therefore requires $\alpha_2$, $\beta_2$, and $\gamma_2$.
Throughout, the weights for the interior residuals were fixed at $\alpha_1=\alpha_2=1$, and the boundary-loss coefficient was set to $c\in\{1,10,100\}$.
Under these settings, TRPINN was configured to satisfy $\beta_2+\gamma_2=c$, and multiple variants were tested for each case.
\tablename~\ref{V=20_rate1} reports the numerical results for $c=1$, \tablename~\ref{V=20_rate2} for $c=10$, and \tablename~\ref{V=20_rate3} for $c=100$.
\tablename~\ref{V=20_ms} summarizes experiments conducted across a range of model sizes.
In these experiments, the vanilla PINN uses the weight [1,100], whereas TRPINN uses [1,50,50].

\begin{figure}[H]
    \centering
    \includegraphics[width=0.8\textwidth]{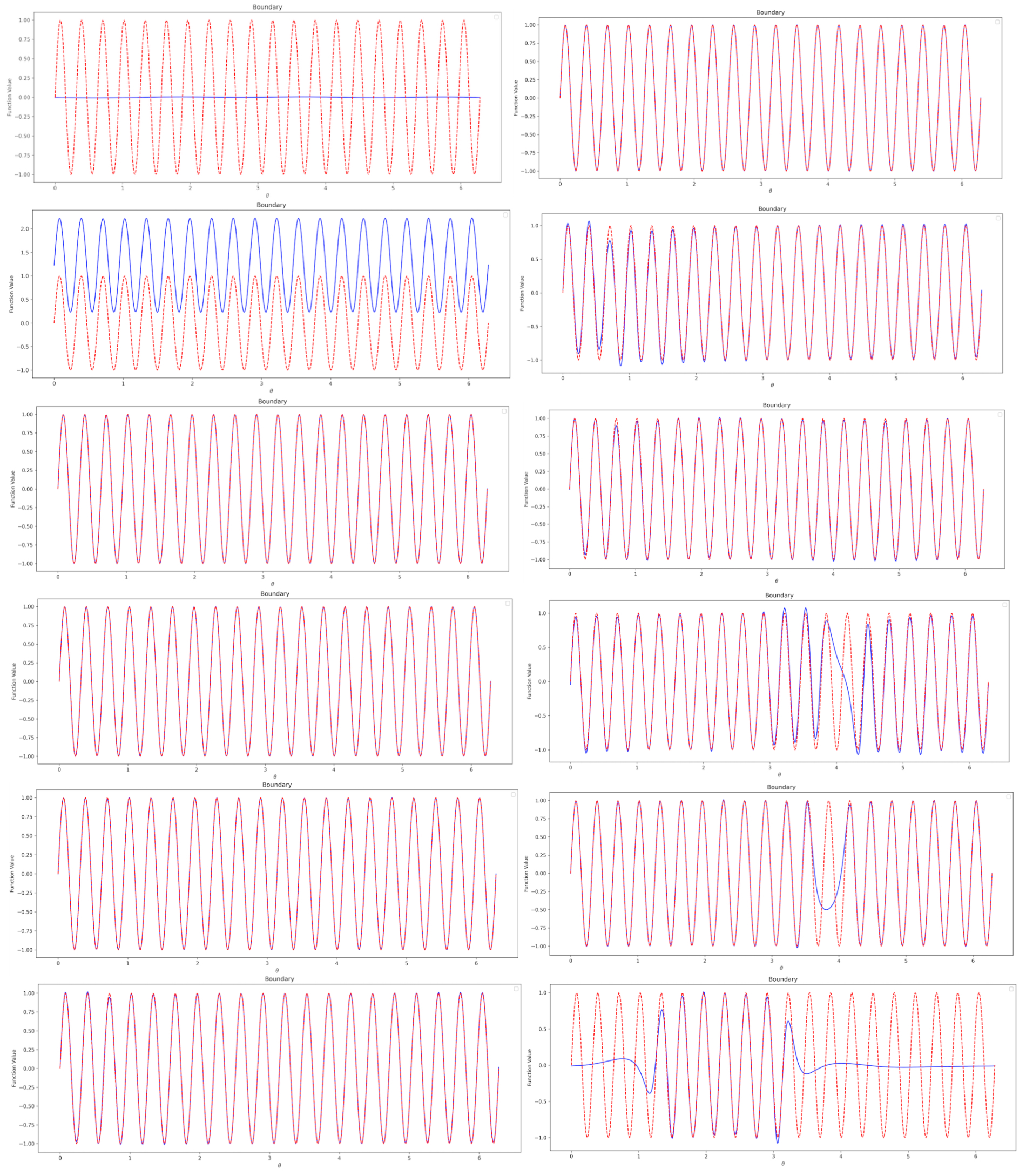}
%\end{figure}
%\begin{figure}[H]
%    \centering
%    \includegraphics[width=0.8\textwidth]{20_rate_bd2}
    \caption{Top-left panel shows the boundary prediction of the vanilla PINN with weights \([1,1]\); all remaining panels show TRPINN. The weights follow the same order as the panels—from the top-left to the bottom-right—as listed in \tablename\ \ref{V=20_rate1}.} 
    \label{20_rate_bd} 
\end{figure}

For the vanilla PINN, all relative errors are approximately $1$, indicating convergence to a trivial solution and, effectively, an absence of learning.
In contrast, TRPINN reports values at the first and third decimal places.
This suggests that even a slight enforcement of the semi-norm leads to improved stability and accuracy of the learned solution.
Consequently, TRPINN avoids collapse to the trivial solution and enables a more faithful reconstruction in the interior and on the boundary.
\figurename~\ref{20_rate_bd} presents the boundary predictions (blue solid) and the true boundary values (red dashed).
The panels are ordered to match \tablename~\ref{V=20_rate1} row-by-row from top to bottom, laid out column-wise starting at the \textit{top-left} panel and proceeding downward, then moving rightward, and ending at the \textit{bottom-right} panel.
The top-left panel displays the boundary prediction of the vanilla PINN, which fails to satisfy the prescribed boundary condition and thus does not yield a meaningful approximation.
In contrast, the bottom-left panel shows that imposing only a 1\% semi-norm in the boundary loss already captures part of the solution.
Proceeding upward in that column, as the boundary-loss proportion increases—i.e., as $\gamma_2$ grows—the predictions approach the true solution more closely.
As seen in \tablename~\ref{V=20_rate1}, the improvement over the vanilla PINN reaches up to three decimal places.
The second panel from the top-left corresponds to $[\alpha_2,\beta_2,\gamma_2]=[1,0,1]$, where no boundary $\mathrm{L}^2$ norm is enforced; nevertheless, the boundary shape is reproduced with high fidelity, underscoring the role of the semi-norm loss in capturing the solution's intricate geometry.

\begin{table}[H]
\centering
\caption{Results of $c=1$ and $V=20$}
\label{V=20_rate1}
\begin{adjustbox}{width=0.6\textwidth}
\begin{tabular}{ll rrrr}
\toprule
Model & Weights&
\multicolumn{4}{c}{Relative errors} \\
\cmidrule(lr){3-6}
&& $\mathrm{H}^1$ inside &$\mathrm{L}^2$ inside &$\mathrm{H}^{\frac{1}{2}}$ boundary &$\mathrm{L}^2$ boundary \\
\midrule
vanilla PINN & [1, 1]         &1.0000E+00&1.0001E+00&9.9939E-01&1.0012E+00\\
\midrule
TRPINN    & [1, 0, 1]        &2.4881E-01&1.0113E+01&4.2658E-03&2.1449E+00\\
TRPINN    & [1, 0.1, 0.9]    &4.3953E-03&5.5286E-03&4.0889E-03&3.6219E-03\\
TRPINN    & [1, 0.2, 0.8]    &6.1923E-03&6.4784E-03&5.0505E-03&3.8178E-03\\
TRPINN    & [1, 0.3, 0.7]    &5.5150E-03&4.1738E-03&5.6181E-03&3.6437E-03\\
TRPINN    & [1, 0.4, 0.6]    &1.7989E-02&4.2121E-02&1.6780E-02&1.8753E-02\\
TRPINN    & [1, 0.5, 0.5]    &4.7755E-03&9.1912E-03&3.7428E-03&3.8010E-03\\
TRPINN    & [1, 0.6, 0.4]    &5.1466E-02&1.6318E-01&6.6900E-02&6.6900E-02\\
TRPINN    & [1, 0.7, 0.3]    &2.7919E-02&8.6616E-02&3.0314E-02&3.3109E-02\\
TRPINN    & [1, 0.8, 0.2]    &2.6885E-01&4.9049E-01&1.8213E-01&3.0699E-01\\
TRPINN    & [1, 0.9, 0.1]    &2.8611E-01&3.6417E-01&1.9953E-01&2.8343E-01\\
TRPINN    & [1, 0.99, 0.01]  &8.2125E-01&8.3445E-01&8.8554E-01&8.6473E-01\\
\bottomrule
\end{tabular}
\end{adjustbox}
\end{table}

In \tablename~\ref{V=20_rate2} and \tablename~\ref{V=20_rate3}, the vanilla PINN yields numerically better training results than in \tablename~\ref{V=20_rate1}.
In contrast, TRPINN maintains performance comparable to \tablename~\ref{V=20_rate1} and still exhibits a relative error at the third decimal place.
This result clearly demonstrates that TRPINN facilitates straightforward selection of the weights $[\alpha,\beta,\gamma]$.

\begin{table}[H]
\centering
\caption{Results of $c=10$ and $V=20$}
\label{V=20_rate2}
\begin{adjustbox}{width=0.6\textwidth}
\begin{tabular}{ll rrrr}
\toprule
Model & Weights&
\multicolumn{4}{c}{Relative errors} \\
\cmidrule(lr){3-6}
&& $\mathrm{H}^1$ inside &$\mathrm{L}^2$ inside &$\mathrm{H}^{\frac{1}{2}}$ boundary &$\mathrm{L}^2$ boundary \\
\midrule
vanilla PINN & [1,10]	&7.6304E-01	&7.6810E-01	&7.9511E-01	&7.7361E-01\\
\midrule
TRPINN    & [1, 1, 9]	&2.5050E-03	&2.6304E-03	&1.9062E-03	&1.5721E-03\\
TRPINN    & [1, 2, 8]	&3.5356E-03	&4.4785E-03	&3.0984E-03	&2.2168E-03\\
TRPINN    & [1, 3, 7]	&4.6392E-03	&6.6775E-03	&3.8406E-03	&3.4119E-03\\
TRPINN    & [1, 4, 6]	&2.5081E-03	&4.8703E-03	&2.1158E-03	&2.0088E-03\\
TRPINN    & [1, 5, 5]	&3.2938E-03	&5.3179E-03	&2.5260E-03	&2.5028E-03\\
TRPINN    & [1, 6, 4]	&2.0471E-03	&1.8646E-03	&1.5651E-03	&1.2188E-03\\
TRPINN    & [1, 7, 3]	&3.6454E-03	&3.2598E-03	&3.5613E-03	&2.3645E-03\\
TRPINN    & [1, 8, 2]	&2.1653E-03	&1.4924E-03	&1.9109E-03	&1.3251E-03\\
TRPINN    & [1, 9, 1]	&5.1616E-03	&4.0353E-03	&4.6346E-03	&3.3409E-03\\
TRPINN    & [1, 9.9, 0.1]	&2.7357E-02	&2.0172E-02	&2.6349E-02	&1.8996E-02\\
\bottomrule
\end{tabular}
\end{adjustbox}
\end{table}

\begin{table}[H]
\centering
\caption{Results of $c=100$ and $V=20$}
\label{V=20_rate3}
\begin{adjustbox}{width=0.6\textwidth}
\begin{tabular}{ll rrrr}
\toprule
Model & Weights&
\multicolumn{4}{c}{Relative errors} \\
\cmidrule(lr){3-6}
&& $\mathrm{H}^1$ inside &$\mathrm{L}^2$ inside &$\mathrm{H}^{\frac{1}{2}}$ boundary &$\mathrm{L}^2$ boundary \\
\midrule
vanilla PINN & [1,100]	&1.5135E-02	&8.2616E-03	&1.3250E-02	&9.0016E-03\\
\midrule
TRPINN	&[1, 10, 90]	&6.8441E-03	&1.2340E-02	&3.3424E-03 &2.7838E-03\\
TRPINN	&[1, 20, 80]	&5.6386E-03	&6.8470E-03	&3.2899E-03 &2.2318E-03\\
TRPINN	&[1, 30, 70]	&5.6901E-03	&5.3552E-03	&3.0845E-03 &2.2753E-03\\
TRPINN	&[1, 40, 60]	&3.0313E-03	&3.2160E-03	&2.1354E-03 &1.3353E-03\\
TRPINN	&[1, 50, 50]	&2.0017E-03	&2.3806E-03	&1.0926E-03 &7.4148E-04\\
TRPINN	&[1, 60, 40]	&3.5143E-03	&3.7460E-03	&1.8404E-03 &1.2850E-03\\
TRPINN	&[1, 70, 30]	&3.3235E-03	&4.1894E-03	&2.1645E-03 &1.3074E-03\\
TRPINN	&[1, 80, 20]	&2.9196E-03	&2.4964E-03	&1.9520E-03 &1.3437E-03\\
TRPINN	&[1, 90, 10]	&2.9295E-03	&2.3582E-03	&1.9707E-03 &1.3070E-03\\
TRPINN	&[1, 99, 1]	&3.1864E-03	&1.3454E-03	&2.1092E-03 &1.3780E-03\\
\bottomrule
\end{tabular}
\end{adjustbox}
\end{table}

\begin{figure}[H]
    \centering
    \includegraphics[scale=0.4]{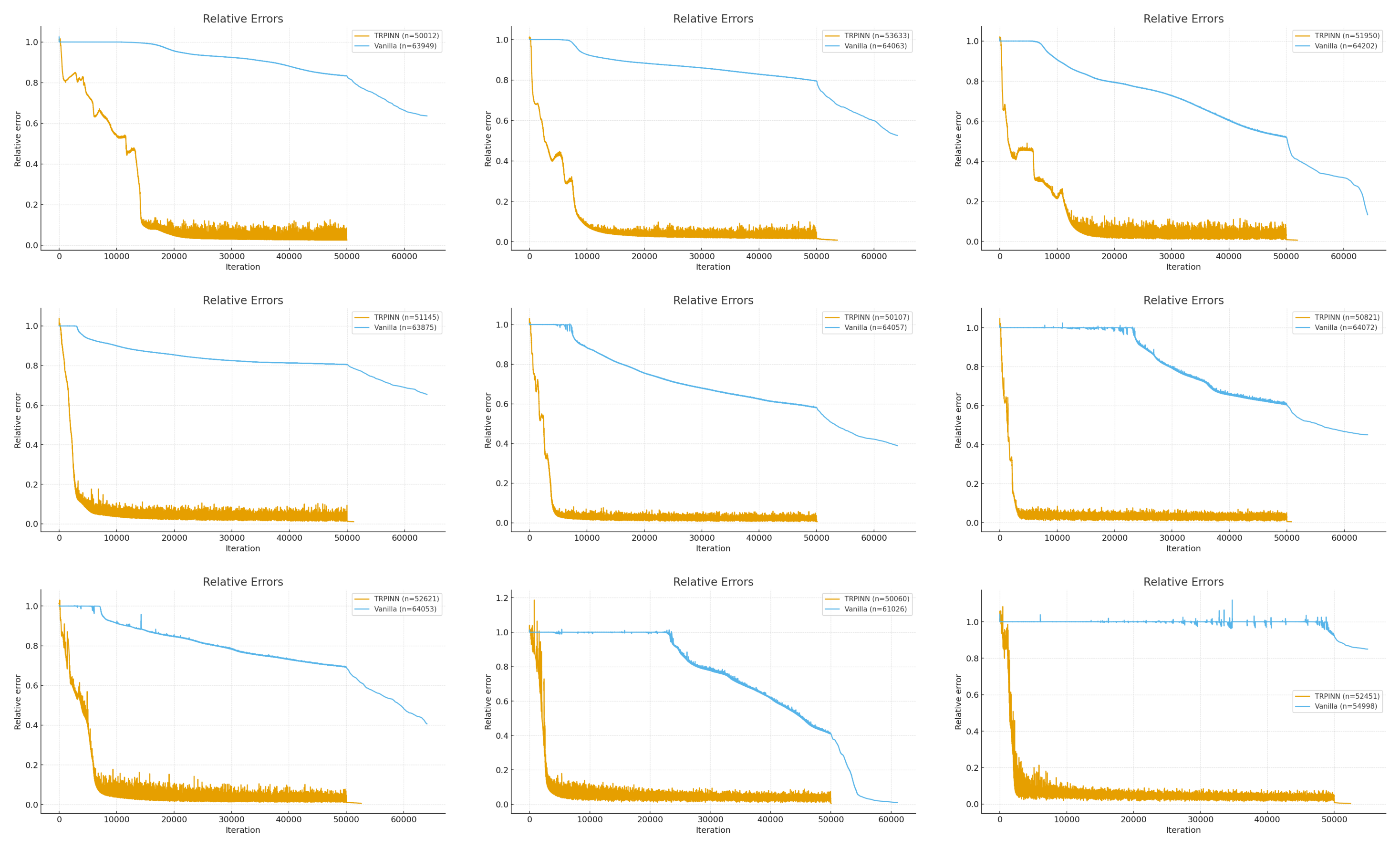} % 이미지 파일 이름
    \caption{\textbf{Relative $\mathrm{H}^1(\Omega$) error}.
    Top row: 3 layers; middle row: 5 layers; bottom row: 7 layers.
    Left column: 30 units; center column: 50 units; right column: 100 units.
    The x-axis denotes iterations: 0--49,999 correspond to Adam training, while iterations $\ge 50,000$ correspond to L-BFGS training.
}
    \label{errors_vis_20_ms} % 참조용 라벨
\end{figure}

Table~\ref{V=20_ms} reports results across a variety of model sizes.
In this configuration, the vanilla PINN adopts the weights [1,100], whereas TRPINN employs [1,50,50].
We observe that, in nearly all cases, the relative errors differ by approximately one to two decimal places.
The table also includes the elapsed times for Adam and L-BFGS.
All Adam runs were trained for 50,000 iterations, and their elapsed times are broadly similar, indicating that the computational cost of vanilla PINN and TRPINN is comparable.
The L-BFGS optimizer is configured to terminate early when further training is not meaningful.
In practice, vanilla PINN typically trained for about 10-20 minutes, while TRPINN incurred negligible additional time during the L-BFGS stage.
Notably, despite the longer training duration, the vanilla PINN underperforms TRPINN.
\figurename~\ref{errors_vis_20_ms} compares the relative $\mathrm{H}^1(\Omega)$ errors of vanilla PINN and TRPINN for each model size in Table~\ref{V=20_ms}.
For PINN, training with Adam shows minimal learning, while L-BFGS yields limited improvement before early stopping.
In contrast, TRPINN exhibits considerable learning with Adam, transitions to L-BFGS for additional improvement, and then reaches early stopping shortly thereafter.

\begin{table}[H]
\centering
\caption{Results by model size of V = 20}
\label{V=20_ms}
\begin{adjustbox}{width=0.9\textwidth}
\begin{tabular}{lll cc rrrr}
\toprule
Model & Weights & Model sizes &
\multicolumn{2}{c}{Elapsed times} &
\multicolumn{4}{c}{Relative errors} \\
\cmidrule(lr){4-5}\cmidrule(lr){6-9}
&&& Adam & LBFGS &
$\mathrm{H}^1$ inside & $\mathrm{L}^2$ inside &
$\mathrm{H}^{\frac{1}{2}}$ boundary & $\mathrm{L}^2$ boundary \\
\midrule
vanilla	&[1, 100]	&30 units/ 3 layers	&00:35:13	&00:10:50	&6.3613E-01	&6.5809E-01	&6.3223E-01	&6.2430E-01\\
TRPINN	&[1, 50, 50]	&30 units/ 3 layers	&00:34:30	&00:00:01	&2.4816E-02	&1.0152E-01	&1.4074E-02	&2.3895E-02\\
vanilla	&[1, 100]	&50 units/ 3 layers	&00:38:21	&00:12:50	&5.2620E-01	&5.4084E-01	&4.5364E-01	&5.0480E-01\\
TRPINN	&[1, 50, 50]	&50 units/ 3 layers	&00:38:12	&00:02:03	&7.9628E-03	&1.0491E-02	&4.7785E-03	&3.5769E-03\\
vanilla	&[1, 100]	&100 units/ 3 layers	&00:46:23	&00:17:44	&1.3280E-01	&1.2274E-01	&1.0677E-01	&1.0464E-01\\
TRPINN	&[1, 50, 50]	&100 units/ 3 layers	&00:47:23	&00:01:31	&5.7988E-03	&7.5700E-03	&3.3798E-03	&2.5857E-03\\
vanilla	&[1, 100]	&30 units/ 5 layers	&00:46:42	&00:13:57	&6.5400E-01	&6.7483E-01	&6.2656E-01	&6.4973E-01\\
TRPINN	&[1, 50, 50]	&30 units/ 5 layers	&00:46:54	&00:00:47	&1.0287E-02	&2.2131E-02	&5.0776E-03	&4.8411E-03\\
vanilla	&[1, 100]	&50 units/ 5 layers	&00:53:07	&00:15:27	&3.8902E-01	&4.1448E-01	&3.4819E-01	&3.7347E-01\\
TRPINN	&[1, 50, 50]	&50 units/ 5 layers	&00:49:35	&00:00:06	&5.5626E-03	&1.1951E-02	&3.5143E-03	&2.6292E-03\\
vanilla	&[1, 100]	&100 units/ 5 layers	&01:09:51	&00:25:33	&4.5067E-01	&4.7581E-01	&4.3940E-01	&4.4373E-01\\
TRPINN	&[1, 50, 50]	&100 units/ 5 layers	&01:09:48	&00:01:13	&4.8237E-03	&4.2851E-03	&2.3875E-03	&1.7449E-03\\
vanilla	&[1, 100]	&30 units/ 7 layers	&01:01:27	&00:16:51	&4.0656E-01	&4.4045E-01	&4.3317E-01	&4.0932E-01\\
TRPINN	&[1, 50, 50]	&30 units/ 7 layers	&01:00:18	&00:02:01	&5.9882E-03	&2.9205E-02	&2.4384E-03	&1.9501E-03\\
vanilla	&[1, 100]	&50 units/ 7 layers	&01:04:28	&00:15:13	&1.1031E-02	&1.2896E-02	&9.8956E-03	&7.7044E-03\\
TRPINN	&[1, 50, 50]	&50 units/ 7 layers	&01:04:08	&00:00:06	&5.8958E-03	&1.7516E-02	&2.5199E-03	&2.2817E-03\\
vanilla	&[1, 100]	&100 units/ 7 layers	&01:35:00	&00:12:16	&8.5004E-01	&8.6021E-01	&8.4430E-01	&8.3441E-01\\
TRPINN	&[1, 50, 50]	&100 units/ 7 layers	&01:34:45	&00:04:51	&3.6996E-03	&3.4170E-03	&1.8699E-03	&1.3352E-03\\
\bottomrule
\end{tabular}
\end{adjustbox}
\end{table}

\subsection{Boundary Condition with Multiple Sharp Peaks}
\begin{figure}[H]
    \centering
    \includegraphics[scale=0.25]{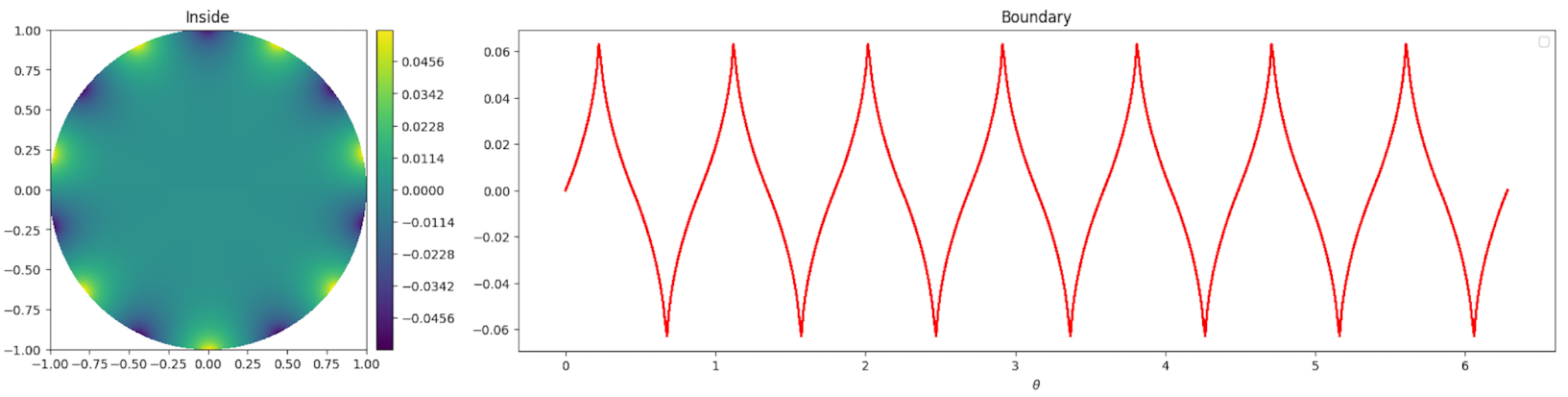}
    \caption{Exact solution: Harmonic extension of $g$ in the section 7.2.} % 캡션
    \label{VAsol} % 참조용 라벨
\end{figure}

In this section, we conduct experiments using $g(t)$ specified in \eqref{VA}.
We state the precise form of the periodic function $g_1$ given on $[0,2\pi]$ in \eqref{VA}:
\begin{align*}
  &g_1(t) = \begin{cases}
    \frac{1}{2} - \sqrt{|\frac{t - \pi/2}{2\pi}|} \quad \quad 0 \le t < \pi,\\
    \sqrt{|\frac{t - 3\pi/2}{2\pi}|} - \frac{1}{2} \quad \quad \pi \le t \le 2\pi,
  \end{cases}\\[1em]
  &g(t) = \frac{1}{V}g_1(Vt) \quad \quad 0 \le t \le 2\pi.
\end{align*}
Although $g$ is defined on $[0,2\pi]$, it can be regarded as a function on the boundary of the unit ball, $\partial B$.
Consequently, it may be prescribed as the boundary condition in \eqref{exam-lap}.
The ground-truth solution was likewise obtained via harmonic extension, and the relative error was computed using it.
\figurename~\ref{VAsol} visualizes the ground-truth solution in the interior (top) and on the boundary (bottom).
We set $V=7$, while keeping all other hyperparameters identical to the previous configuration.
The weights are listed in \tablename~\ref{7VA}.
The table shows that the vanilla PINN yields relative errors near $1$ in all cases, indicating complete training failure under the prescribed settings.
In contrast, the proposed TRPINN achieves improvements of two decimal places in the relative $\mathrm{L}^2$ error and one decimal place in the relative $\mathrm{H}^1$ error.
Moreover, the results are broadly consistent across different weights choices, suggesting that stable convergence can be obtained without extensive tuning of the weights.

\begin{figure}[H]
    \centering
    \includegraphics[width=0.8\textwidth]{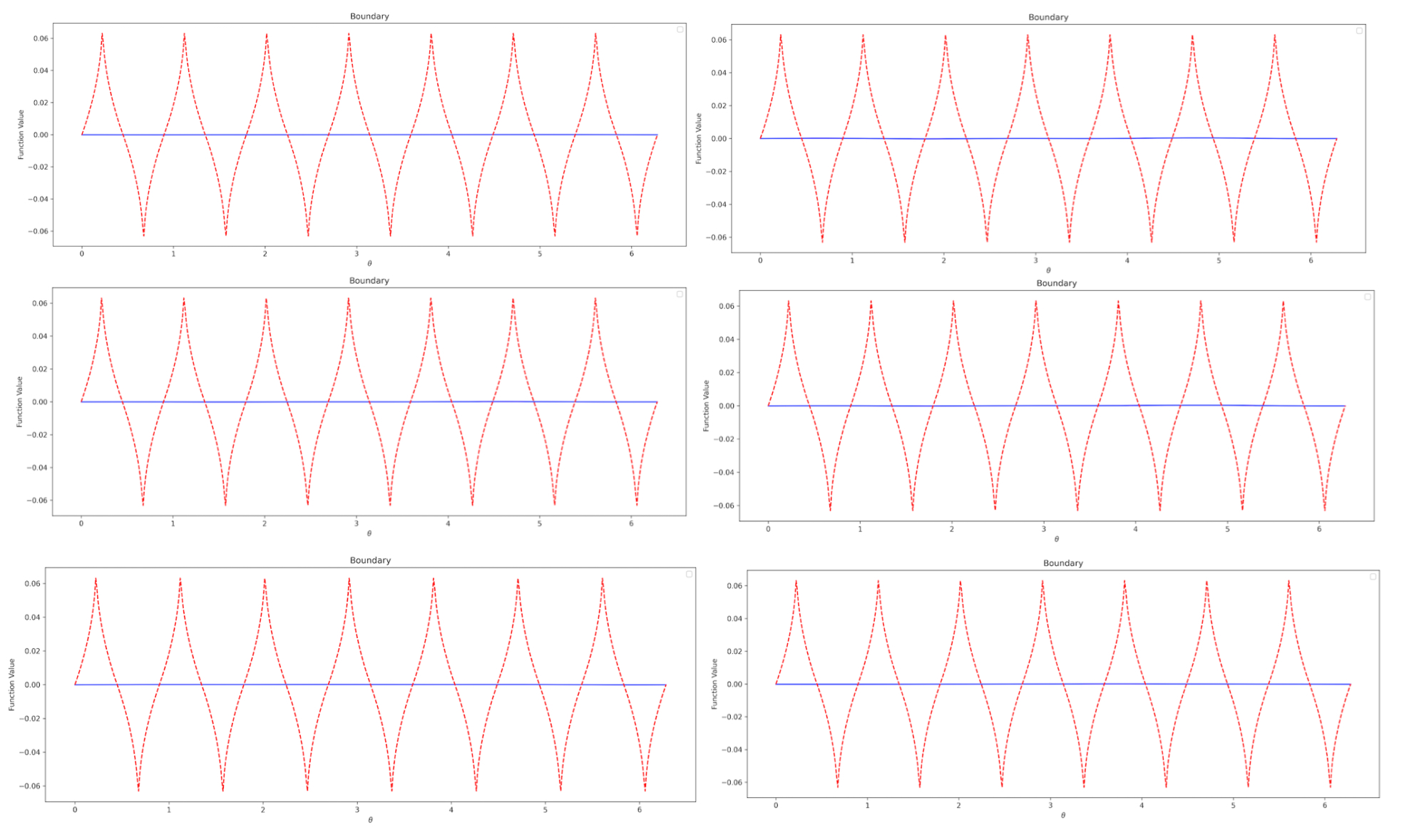} % 이미지 파일 이름
    \caption{These panels show the boundary prediction of the vanilla PINN with six weights in \tablename~\ref{7VA}} % 캡션
    \label{7VAVA} % 참조용 라벨
\end{figure}

\begin{figure}[H]
    \centering
    \includegraphics[width=0.8\textwidth]{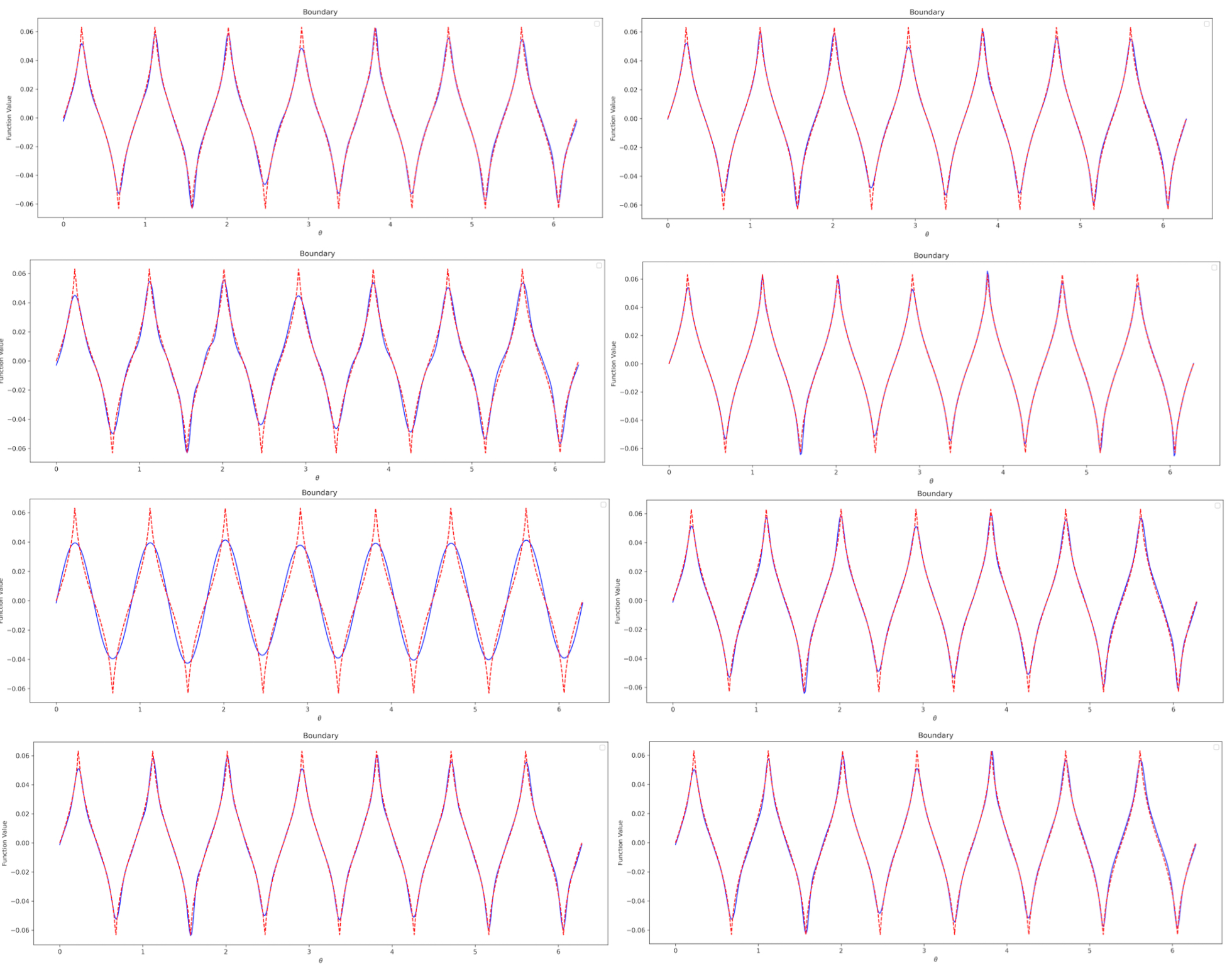} % 이미지 파일 이름
    \caption{These panels show the boundary prediction of TRPINN with eight weights in \tablename~\ref{7VA}} % 캡션
    \label{7VAsemi} % 참조용 라벨
\end{figure}

\figurename~\ref{7VAVA} presents visualizations of the vanilla PINN across the tested weights, while \figurename~\ref{7VAsemi} displays the boundary visualizations for TRPINN.
As evident from \figurename~\ref{7VAVA}, the vanilla PINN fails under all specified weights, producing outputs that are nearly zero.
In contrast, TRPINN yields boundary predictions that are nearly accurate in all cases.

\begin{table}[H]
\centering
\caption{Results of boundary condition with sharp peaks}
\label{7VA}
\begin{adjustbox}{width=0.6\textwidth}
\begin{tabular}{llrrrr}
\toprule
Model & Weights&
\multicolumn{4}{c}{Relative errors} \\
\cmidrule(lr){3-6}
&& $\mathrm{H}^1$ inside &$\mathrm{L}^2$ inside &$\mathrm{H}^{\frac{1}{2}}$ boundary &$\mathrm{L}^2$ boundary \\
\midrule
vanilla PINN & [0.1, 10]          &1.0000E+00 &9.9999E-01 &9.9919E-01 &9.9086E-01\\
vanilla PINN & [1, 1]             &9.9086E-01 &9.9999E-01 &9.9908E-01 &9.8782E-01\\
vanilla PINN & [1, 10]            &1.0000E+00 &1.0000E+00 &9.9851E-01 &9.8876E-01\\
vanilla PINN & [1, 100]           &1.0000E+00 &1.0000E+00 &9.9671E-01 &9.8397E-01\\
vanilla PINN & [1, 1000]          &1.0000E+00 &1.0000E+00 &9.9835E-01 &1.0020E+00\\
vanilla PINN & [1, 10000]         &1.0000E+00 &1.0000E+00 &9.9916E-01 &1.0053E+00\\
\midrule
TRPINN    & [0.1, 10, 1]       &1.5017E-01 &5.2118E-02 &2.1964E-01 &7.3685E-02\\
TRPINN    & [1, 1, 1]          &2.3112E-01 &9.9885E-02 &2.9822E-01 &1.4324E-01\\
TRPINN    & [1, 10, 1]         &4.4464E-01 &1.6449E-01 &4.6405E-01 &2.5043E-01\\
TRPINN    & [1, 100, 10]       &1.3471E-01 &5.1674E-02 &2.4676E-01 &9.4027E-02\\
TRPINN    & [1, 1000, 100]     &1.5464E-01 &6.0658E-02 &2.1056E-01 &6.2470E-02\\
TRPINN    & [1, 10000, 1000]   &2.1913E-01 &7.1406E-02 &1.9918E-01 &1.2238E-01\\
TRPINN    & [1, 10, 10]        &1.4415E-01 &6.4043E-02 &2.2320E-01 &8.8819E-02\\
TRPINN    & [1, 100, 100]      &1.5398E-01 &9.2327E-02 &2.0581E-01 &7.0024E-02\\
\bottomrule
\end{tabular}
\end{adjustbox}
\end{table}

\section{discussion}
\label{discussion}

The key finding of this study is that our proposed TRPINN achieves improvements in the $\mathrm{H}^1(\Omega)$ error by one to two orders of magnitude over the vanilla PINN in problems with challenging boundary conditions.
This performance gain can be explained by mechanisms that enforce the theoretically consistent boundary regularity and capture the oscillation and sharpness of the boundary conditions.

In the vanilla PINN, for convenience, the boundary loss typically applies $\mathrm{L}^2(\partial \Omega)$, which may fail to adequately reflect the correct boundary regularity.
In contrast, we design TRPINN by applying $\mathrm{H}^{\frac{1}{2}}(\partial \Omega)$ to the boundary loss, which aligns exactly with PDE theory.
By computing only the essential part of the $\mathrm{H}^{\frac{1}{2}}(\partial \Omega)$ norm and appropriately discretizing it, we improve performance while maintaining a computational cost comparable to that of the vanilla PINN.
For example, when the boundary condition exhibits significant oscillations, the relative error is ahead of the vanilla PINN by one to two decimal places; and for sharp boundary conditions, the boundary predictions of the vanilla PINN can be completely smeared out, whereas TRPINN captures the sharpness of the boundary condition well, as shown in the numerical examples.
Although TRPINN strengthens boundary regularity, it does not require additional information such as the derivative $Dg$ of the boundary condition and, as with the vanilla PINN, it can rely solely on data for the boundary condition $g$.
Moreover, in the vanilla PINN, model performance is highly sensitive to the specific setting of the interior and boundary loss weights \(\alpha\) and \(\beta\), which makes choosing them challenging.
In contrast, TRPINN allows \(\alpha, \beta, \gamma\) to be chosen much more robustly, as demonstrated in the numerical examples.
%\begin{figure}[H]
%    \centering
%    \begin{minipage}{0.48\textwidth}
%        \centering
%        \includegraphics[width=\textwidth]{7VAVA}
%        \caption{These panels show the boundary prediction of the vanilla PINN with six weights in \tablename~\ref{7VA}}
%        \label{7VAVA}
%    \end{minipage}
%    \hfill
%    \begin{minipage}{0.48\textwidth}
%        \centering
%        \includegraphics[width=\textwidth]{7VAsemi}
%        \caption{These panels show the boundary prediction of the TRPINN with eight weights in \tablename~\ref{7VA}}
%        \label{7VAsemi}
%    \end{minipage}
%\end{figure}

We show that, to achieve convergence to the exact solution in $\mathrm{H}^1(\Omega)$, applying the $\mathrm{H}^{\frac{1}{2}}(\partial \Omega)$ norm to the boundary loss is essential.
Theoretically, even if the loss of the vanilla PINN converges well to 0, convergence to the exact solution in $\mathrm{H}^1(\Omega)$ cannot be guaranteed.
For TRPINN, as long as the loss function converges well to 0, the approximate solution converges to the exact solution in $\mathrm{H}^1(\Omega)$.
Indeed, applying $\mathrm{H}^{\frac{1}{2}}(\partial \Omega)$ to the boundary loss is not excessive but rather an appropriate choice from a theoretical standpoint.

\section{conclusion}
\label{conclusion}

In this work, we propose TRPINN, which (i) selectively enforces only the essential components of the $\mathrm{H}^{\frac{1}{2}}(\partial \Omega)$ norm at the boundary with almost no increase in computational cost for the boundary loss, (ii) substantially reduces the $\mathrm{H}^1(\Omega)$ relative error, (iii) theoretically guarantees that if the loss function converges to zero, the approximate solution converges in $\mathrm{H}^1(\Omega)$, (iv) accelerates convergence by mitigating the per-mode learning-speed gap from the NTK perspective, and (v) makes performance more \emph{robust} to the choice of boundary-loss weights.
These results indicate that, when the $\mathrm{H}^{\frac{1}{2}}(\partial \Omega)$ constraint is aligned with PDE theory, it supports the reduction of $\mathrm{H}^1(\Omega)$ error, and they suggest tangible performance gains on real problems with complicated boundary conditions.
Finally, the applicability of TRPINN can be broadened by extending it to sources $f \in \mathrm{H}^{-1}(\Omega)$ and to non-stationary (time-dependent) equations.

\section*{Acknowledgements}
We thank Donghun Lee for helpful comments on the manuscript.
We are also grateful to Jeongun Ha for advice on the experimental setup.
Finally, we acknowledge Suyoung Min for substantial assistance with the early experiments conducted prior to drafting this paper.

%\newpage
\appendix

\section{Approximation solutions using approximate coefficients}

To obtain an approximation solution, i.e., a neural network $u_{NN}$ to the equation \eqref{divergence-form} using PINNs, it is required that one has $\sum_{i,j=1}^d D_i(a^{ij}D_ju_{NN}) \in \mathrm{L}^2(\Omega)$ and $f \in \mathrm{L}^2(\Omega)$ (this means that each $f_i$ = 0).
However, in the divergence form, \(\mathcal{L}\) does not, in general (especially with no regularity on $a^{ij}$), map into \(\mathrm{L}^2(\Omega)\).
Therefore, theoretical justifications are in order to accommodate divergence-form PDEs within the PINN framework.

Assume that all \( f_i \) are zero and that the coefficients \( a^{ij} \) are in $\mathrm{W}_\infty^1(\Omega)$.
In this case,
\[
f_{NN}: = \sum_{i,j=1}^d D_i(a^{ij}D_ju_{NN})\in \mathrm{L}^2(\Omega),
\]
so one can directly apply PINN.
For general $a^{ij} \in \mathrm{L}^{\infty}(\Omega)$ with no regularity and $f_i = 0$, one can proceed as follows to obtain approximation $u_{NN}$ which is sufficiently close to the true solution $u$.

\begin{lem}
  For the coefficients $a^{ij} \in \mathrm{L}^{\infty}(\Omega)$, we have $a^{ij}_{\varepsilon}\in \mathrm{W}_\infty^1(\Omega)$ such that
\begin{align*}
&\text{(1)} \quad a^{ij}_{\varepsilon}(x) \rightarrow a^{ij}(x) \text{ as } \varepsilon \rightarrow 0 \text{ for a.e.}\; x\in \Omega.
\\
&\text{(2)} \quad a^{ij}_{\varepsilon} \text{ satisfy the same ellipticity as }a^{ij}.
\\
&\text{(3)} \quad a^{ij}_{\varepsilon}, Da^{ij}_{\varepsilon} \text{ are bounded on }\Omega \,\, \text{(the bounds depend on $\varepsilon$)}.
\end{align*}
\end{lem}
The proof follows readily from the standard approximation argument.
Define
\[
\mathcal{L}_{\varepsilon}u := \sum_{i,j=1}^dD_j(a^{ij}_{\varepsilon}D_i u).
\]
There exists a unique solution $u_\varepsilon \in \mathrm{H}^1(\Omega)$ to the following problem:
\begin{align}
  \label{L-ep}
  \begin{cases}
    \mathcal{L}_{\varepsilon}u_\varepsilon = f \text{ in } \Omega\\
    u_\varepsilon|_{\partial \Omega} = g \text{ on } \partial \Omega.
  \end{cases}
\end{align}
We will choose \( \varepsilon \) sufficiently small and apply this problem \eqref{L-ep} in PINN and TRPINN.
For this purpose, the solution to \eqref{L-ep} must be sufficiently close to the true solution $u$ to the problem \eqref{divergence-form}.
The following lemma shows that the solution of \eqref{L-ep} is very close to the solution of the original problem \eqref{divergence-form} in the \( \mathrm{H}^1(\Omega) \) sense.
\begin{lem}
  Let $u$, $u_{\varepsilon} \in \mathrm{H}^1(\Omega)$  be solutions to \eqref{divergence-form} and \eqref{L-ep}, respectively, with the same forcing function \(f\) and boundary data \(g\).
  Then 
  \begin{align}
  \label{ep-LDCT}
    \rVert u - u_{\varepsilon} \lVert_{\mathrm{H}^1(\Omega)} \le C \rVert(a^{ij} - a^{ij}_{\varepsilon})Du \lVert_{\mathrm{L}^2(\Omega)},
  \end{align}
  where $C$ is independent of $\varepsilon$.
  (This lemma remains valid even when \( f_i \neq 0 \).)
\end{lem}
Letting $\varepsilon \to 0$, the LDCT implies that the left-hand side of \eqref{ep-LDCT} converges to zero.
Consequently, even when $a^{ij}$ belongs only to $\mathrm{L}^{\infty}(\Omega)$, we may choose $\epsilon>0$ sufficiently small and solve \eqref{L-ep} instead, thereby obtaining an approximate solution.

\bibliographystyle{plain}

\bibliography{myrefs.bib}

@book{krylov2024lectures,
  title={Lectures on elliptic and parabolic equations in Sobolev spaces},
  author={Krylov, Nikola{\u\i} Vladimirovich},
  volume={96},
  year={2024},
  publisher={American Mathematical Society}
}

@book {MR503903,
    AUTHOR = {Triebel, Hans},
     TITLE = {Interpolation theory, function spaces, differential operators},
    SERIES = {North-Holland Mathematical Library},
    VOLUME = {18},
 PUBLISHER = {North-Holland Publishing Co., Amsterdam-New York},
      YEAR = {1978},
     PAGES = {528},
      ISBN = {0-7204-0710-9},
   MRCLASS = {46E35 (35Jxx 46M35)},
  MRNUMBER = {503903},
MRREVIEWER = {Robert D. Brown},
}

@book{evans2022partial,
  title={Partial differential equations},
  author={Evans, Lawrence C},
  volume={19},
  year={2022},
  publisher={American mathematical society}
}

@article {MR2352844,
    AUTHOR = {Kim, Doyoon},
     TITLE = {Trace theorems for {S}obolev-{S}lobodeckij spaces with or
              without weights},
   JOURNAL = {J. Funct. Spaces Appl.},
  FJOURNAL = {Journal of Function Spaces and Applications},
    VOLUME = {5},
      YEAR = {2007},
    NUMBER = {3},
     PAGES = {243--268},
      ISSN = {0972-6802},
   MRCLASS = {46E35},
  MRNUMBER = {2352844},
MRREVIEWER = {Dorothee\ D.\ Haroske},
       DOI = {10.1155/2007/471535},
       URL = {https://doi.org/10.1155/2007/471535},
}

@article{ZANG2020109409,
title = {Weak adversarial networks for high-dimensional partial differential equations},
journal = {Journal of Computational Physics},
volume = {411},
pages = {109409},
year = {2020},
issn = {0021-9991},
doi = {https://doi.org/10.1016/j.jcp.2020.109409},
url = {https://www.sciencedirect.com/science/article/pii/S0021999120301832},
author = {Yaohua Zang and Gang Bao and Xiaojing Ye and Haomin Zhou}
}

@article{zhou2025ssbepinnsobolevboundaryscheme,
      title={{SSBE}-{PINN}: {A} {S}obolev Boundary Scheme Boosting Stability and Accuracy in Elliptic/Parabolic {PDE} Learning}, 
      author={Qixuan Zhou and Chuqi Chen and Tao Luo and Yang Xiang},
      journal={arXiv preprint arXiv:2508.10322},
      year = {2025},
      eprint={2508.10322},
      archivePrefix={arXiv},
      primaryClass={math.NA},
      url={https://arxiv.org/abs/2508.10322}
}

@article{liu2023deep,
  title={Deep {R}itz method with adaptive quadrature for linear elasticity},
  author={Liu, Min and Cai, Zhiqiang and Ramani, Karthik},
  journal={Computer Methods in Applied Mechanics and Engineering},
  volume={415},
  pages={116229},
  year={2023},
  publisher={Elsevier}
}

@article{jacot2018neural,
  title={Neural tangent kernel: Convergence and generalization in neural networks},
  author={Jacot, Arthur and Gabriel, Franck and Hongler, Cl{\'e}ment},
  journal={Advances in neural information processing systems},
  volume={31},
  year={2018}
}

@article{carvalho2025positivity,
  title={The positivity of the neural tangent kernel},
  author={Carvalho, Lu{\'\i}s and Costa, Jo{\~a}o L and Mour{\~a}o, Jos{\'e} and Oliveira, Gon{\c{c}}alo},
  journal={SIAM Journal on Mathematics of Data Science},
  volume={7},
  number={2},
  pages={495--515},
  year={2025},
  publisher={SIAM}
}

@article{wang2022and,
  title={When and why PINNs fail to train: A neural tangent kernel perspective},
  author={Wang, Sifan and Yu, Xinling and Perdikaris, Paris},
  journal={Journal of Computational Physics},
  volume={449},
  pages={110768},
  year={2022},
  publisher={Elsevier}
}

@inproceedings{rahaman2019spectral,
  title={On the spectral bias of neural networks},
  author={Rahaman, Nasim and Baratin, Aristide and Arpit, Devansh and Draxler, Felix and Lin, Min and Hamprecht, Fred and Bengio, Yoshua and Courville, Aaron},
  booktitle={International conference on machine learning},
  pages={5301--5310},
  year={2019},
  organization={PMLR}
}

@article{cao2019towards,
  title={Towards understanding the spectral bias of deep learning},
  author={Cao, Yuan and Fang, Zhiying and Wu, Yue and Zhou, Ding-Xuan and Gu, Quanquan},
  journal={arXiv preprint arXiv:1912.01198},
  year={2019}
}

@article{tancik2020fourier,
  title={Fourier features let networks learn high frequency functions in low dimensional domains},
  author={Tancik, Matthew and Srinivasan, Pratul and Mildenhall, Ben and Fridovich-Keil, Sara and Raghavan, Nithin and Singhal, Utkarsh and Ramamoorthi, Ravi and Barron, Jonathan and Ng, Ren},
  journal={Advances in neural information processing systems},
  volume={33},
  pages={7537--7547},
  year={2020}
}

@inproceedings{basri2020frequency,
  title={Frequency bias in neural networks for input of non-uniform density},
  author={Basri, Ronen and Galun, Meirav and Geifman, Amnon and Jacobs, David and Kasten, Yoni and Kritchman, Shira},
  booktitle={International conference on machine learning},
  pages={685--694},
  year={2020},
  organization={PMLR}
}

@article{raissi2017physics,
  title={Physics informed deep learning (part i): Data-driven solutions of nonlinear partial differential equations},
  author={Raissi, Maziar and Perdikaris, Paris and Karniadakis, George Em},
  journal={arXiv preprint arXiv:1711.10561},
  year={2017}
}

@article{raissi2019physics,
  title={Physics-informed neural networks: A deep learning framework for solving forward and inverse problems involving nonlinear partial differential equations},
  author={Raissi, Maziar and Perdikaris, Paris and Karniadakis, George E},
  journal={Journal of Computational physics},
  volume={378},
  pages={686--707},
  year={2019},
  publisher={Elsevier}
}

@article{toscano2025pinns,
  title={From {PINN}s to {PIKAN}s: Recent advances in physics-informed machine learning},
  author={Toscano, Juan Diego and Oommen, Vivek and Varghese, Alan John and Zou, Zongren and Ahmadi Daryakenari, Nazanin and Wu, Chenxi and Karniadakis, George Em},
  journal={Machine Learning for Computational Science and Engineering},
  volume={1},
  number={1},
  pages={1--43},
  year={2025},
  publisher={Springer}
}

@article{kharazmi2019variational,
  title={Variational physics-informed neural networks for solving partial differential equations},
  author={Kharazmi, Ehsan and Zhang, Zhongqiang and Karniadakis, George Em},
  journal={arXiv preprint arXiv:1912.00873},
  year={2019}
}

@article{lu2021physics,
  title={Physics-informed neural networks with hard constraints for inverse design},
  author={Lu, Lu and Pestourie, Raphael and Yao, Wenjie and Wang, Zhicheng and Verdugo, Francesc and Johnson, Steven G},
  journal={SIAM Journal on Scientific Computing},
  volume={43},
  number={6},
  pages={B1105--B1132},
  year={2021},
  publisher={SIAM}
}

@article{mcclenny2023self,
  title={Self-adaptive physics-informed neural networks},
  author={McClenny, Levi D and Braga-Neto, Ulisses M},
  journal={Journal of Computational Physics},
  volume={474},
  pages={111722},
  year={2023},
  publisher={Elsevier}
}

@article{wang2024improved,
  title={An improved physics-informed neural network with adaptive weighting and mixed differentiation for solving the incompressible Navier--Stokes equations},
  author={Wang, Jie and Xiao, Xufeng and Feng, Xinlong and Xu, Hui},
  journal={Nonlinear Dynamics},
  volume={112},
  number={18},
  pages={16113--16134},
  year={2024},
  publisher={Springer}
}

@article{wu2022randomized,
  title={A randomized trapezoidal quadrature},
  author={Wu, Yue},
  journal={International Journal of Computer Mathematics},
  volume={99},
  number={4},
  pages={680--692},
  year={2022},
  publisher={Taylor \& Francis}
}

\end{document}